\documentclass[journal]{IEEEtran}
\usepackage{amsmath,amsfonts}
\usepackage{algorithm}
\usepackage{array}
\usepackage[caption=false,font=normalsize,labelfont=sf,textfont=sf]{subfig}
\usepackage{textcomp}
\usepackage{stfloats}
\usepackage{bm}
\usepackage{url}
\usepackage{verbatim}
\usepackage{graphicx}
\usepackage{xcolor}
\usepackage{hyperref}
\usepackage{cleveref}
\usepackage{cite}
\usepackage{algpseudocode}
\usepackage{booktabs}  
\usepackage{xcolor}
\newcounter{eqtagb}
\newcounter{eqtagc}
\newcounter{eqtagd}

\begin{document}

\title{Collision Detection with Analytical Derivatives of Contact Kinematics}

\author{Anup~Teejo~Mathew\textsuperscript{1,2}, 
        Anees~Peringal\textsuperscript{1},
        Daniele~Caradonna\textsuperscript{3,4},
        Frederic~Boyer\textsuperscript{5}, 
        and~Federico~Renda\textsuperscript{1,2}%
\thanks{\textsuperscript{1}Department of Mechanical and Nuclear Engineering, Khalifa University of Science and Technology, Abu Dhabi, UAE.}%
\thanks{\textsuperscript{2}Khalifa University Center for Autonomous Robotic Systems (KUCARS), Abu Dhabi, UAE.}%
\thanks{\textsuperscript{3}The BioRobotics Institute, Scuola Superiore Sant'Anna, Pisa, Italy.}%
\thanks{\textsuperscript{4}Department of Excellence in Robotics and AI, Scuola Superiore Sant'Anna, Pisa, Italy.}%
\thanks{\textsuperscript{5}LS2N Laboratory, Institut Mines Telecom Atlantique, Nantes 44307, France.}%
\thanks{Corresponding author: Anup Teejo Mathew (email: anup.mathew@ku.ac.ae, anupteejo@gmail.com).}}



\maketitle

\begin{abstract}
Differentiable contact kinematics are essential for gradient-based methods in robotics, yet the mapping from robot state to contact distance, location, and normal becomes non-smooth in degenerate configurations of shapes with zero or undefined curvature. We address this inherent limitation by selectively regularizing such geometries into strictly convex implicit representations, restoring uniqueness and smoothness of the contact map. Leveraging this geometric regularization, we develop iDCOL, an implicit differentiable collision detection and contact kinematics framework. iDCOL represents colliding bodies using strictly convex implicit surfaces and computes collision detection and contact kinematics by solving a fixed-size nonlinear system derived from a geometric scaling-based convex optimization formulation. By applying the Implicit Function Theorem to the resulting system residual, we derive analytical derivatives of the contact kinematic quantities. We develop a fast Newton-based solver for iDCOL and provide an open-source C++ implementation of the framework. The robustness of the approach is evaluated through extensive collision simulations and benchmarking, and applicability is demonstrated in gradient-based kinematic path planning and differentiable contact physics, including multi-body rigid collisions and a soft-robot interaction example.
\end{abstract}


\section{Introduction}
Fast and accurate computation of derivatives of the governing equations of rigid-body dynamics with respect to robot state and control inputs has become a central requirement in modern robotics  \cite{ModernRobotics, Tassa2014DDP}. This capability enables gradient-based methods in planning~\cite{toussaint2018differentiable}, control~\cite{bacher2021design}, learning~\cite{heess2015learning}, and simulation~\cite{carpentier2019pinocchio}. As a result, increasing attention has been directed toward differentiable frameworks and physics engines \cite{Newbury2024}. Existing physics engines rely on numerical differentiation \cite{Todorov2012MuJoCo}, automatic differentiation (autodiff) \cite{howell2023dojodifferentiablephysicsengine, Giftthaler2017}, or closed-form analytical derivatives \cite{Carpentier2019} to compute Jacobians of the system dynamics. Numerical and autodiff-based approaches yield approximate or computationally expensive Jacobians, whereas analytical differentiation, while challenging to derive and implement, is exact and computationally efficient when formulated correctly \cite{Singh2022}. As robots increasingly interact physically with the environment, differentiable frameworks capable of handling contact-rich interactions become essential \cite{LidecReview2024, Lidec2025}. A prerequisite for differentiable contact physics is a well-defined, fast, and differentiable contact kinematics map.

\begin{figure}[t]
    \centering
    \includegraphics[width=0.9\linewidth]{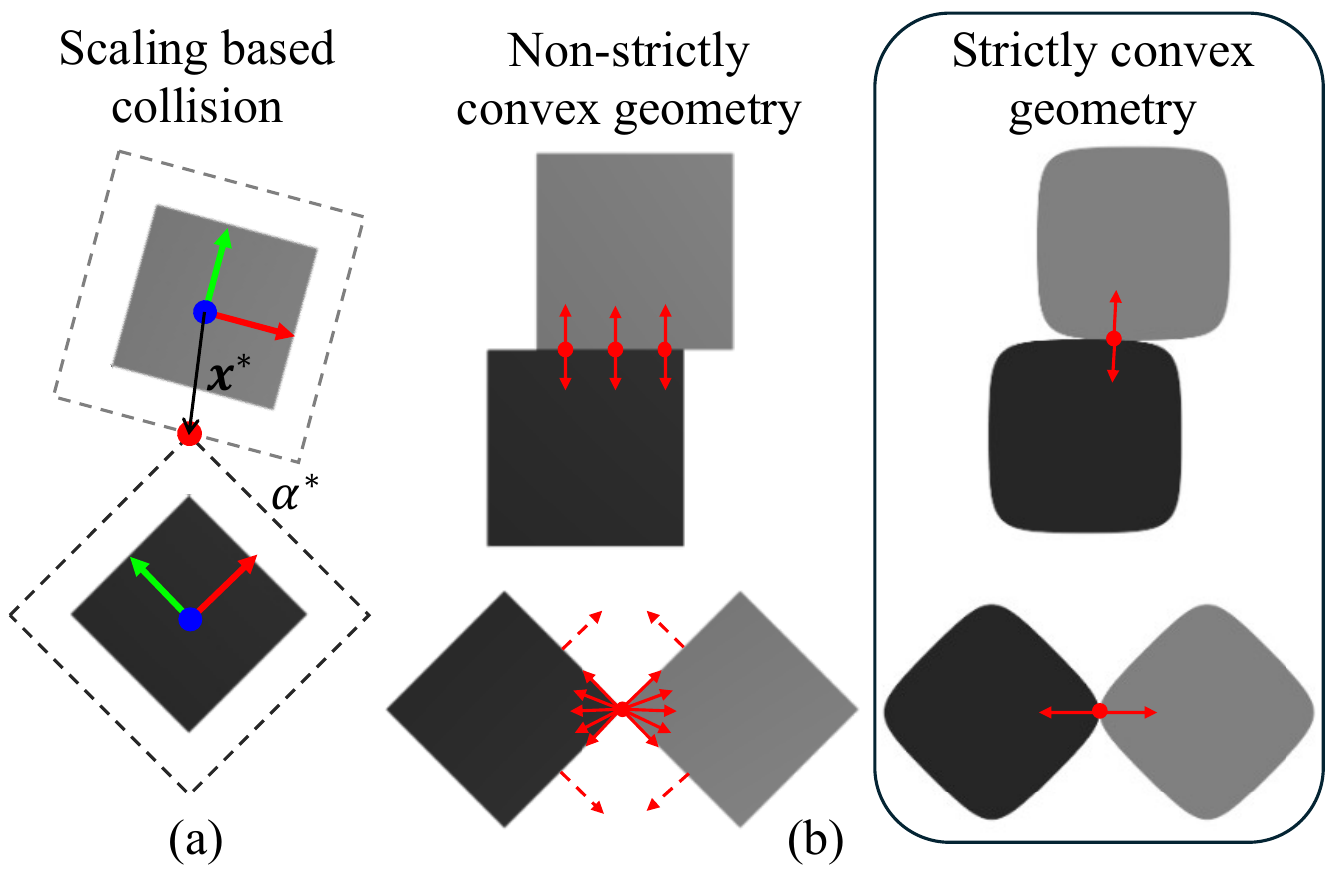}
    \caption{(a) The scaling-based formulation computes the minimum geometric scaling factor $\alpha^*$ for which the scaled convex bodies share a contact point $\bm{x}^*$. $\alpha^*<1$ indicates penetration, $\alpha^*=1$ contact, and $\alpha^*>1$ separation. (b) In non-strictly convex geometries (left), degenerate configurations lead to non-unique contact locations or normals. In contrast, strictly convex geometries (right) admit an isolated contact point and a unique contact normal, yielding well-defined, differentiable contact kinematics.}
    \label{fig:intro}
\end{figure}

Contact kinematics maps robot states to quantities such as contact distance (gap or penetration), contact location, and contact normal. These quantities are typically computed using the Gilbert–Johnson–Keerthi (GJK) algorithm \cite{GJK} for collision detection and distance queries, together with the Expanding Polytope Algorithm (EPA) for recovering penetration depth and contact kinematics in penetrating configurations \cite{vanDenBergen2004}. These methods, widely used in physics engines through libraries such as FCL \cite{FCL}, rely on discrete feature selection and active-set changes, resulting in contact quantities that are piecewise-defined and non-differentiable when contact features switch. Recent work has sought to improve the efficiency and robustness of GJK queries through optimization-based formulations \cite{Montaut2024}, and to enable gradient estimation via randomized smoothing \cite{Montaut2023}. Another approach computes closest distances between geometric primitive pairs by formulating the distance computation as a differentiable optimization problem, enabling analytical gradient computation for a limited set of primitive geometries \cite{Zimmermann2022}.

An alternative class of approaches employs implicit surface representations, most commonly in the form of signed distance fields (SDFs), for collision detection and contact kinematics \cite{SDF, Li2024, Strecke2021}. In these methods, the geometry of a body is represented by an implicit scalar function $\phi:\mathbb{R}^3 \to \mathbb{R}$, whose zero level set defines the surface. Practical implementations typically rely on either sampling-based strategies \cite{Ratliff2009} or optimization-based formulations \cite{Macklin2020, CHEN20061053} that typically minimize a separation distance (gap function) between the bodies. While the latter enable gradient computation, they are generally restricted to separating configurations, with penetration requiring alternative problem formulations \cite{Jaitly2025}.

Recently, Tracy et al. introduced DCOL, a differentiable collision formulation that detects collisions between convex primitives by solving a convex cone program to find the smallest uniform scaling factor for which the two scaled bodies intersect \cite{Tracy2023} (Fig. \ref{fig:intro}(a)). Analytical derivatives of contact kinematics are obtained by applying the Implicit Function Theorem (IFT) to the Karush--Kuhn--Tucker (KKT) residual of the resulting conic program. The application of IFT in DCOL assumes a fixed active constraint set; changes in the active set lead to a piecewise-smooth residual, reflecting underlying geometric degeneracies. 

Beyond algorithmic considerations, differentiability is fundamentally limited by the geometry of the contacting bodies (Fig. \ref{fig:intro}(b)). A geometry is strictly convex if its boundary has positive finite curvature in all principal directions (e.g., spheres and ellipsoids). All other convex geometries are non-strictly convex, exhibiting zero or undefined curvature in at least one direction (e.g., blocks and cylinders). The latter give rise to intrinsic degeneracies in contact kinematics, including non-unique contact locations or normals. In contrast, strictly convex geometries admit unique contact locations and normals, yielding differentiable contact kinematics \cite{Escande2014}. 

Motivated by these observations, we introduce iDCOL, an implicit differentiable contact kinematics framework that regularizes non-strictly convex geometries into
strictly convex implicit representations, restoring differentiability in degenerate contact configurations. Building on the scaling-based perspective of
DCOL, iDCOL represents colliding bodies using smooth implicit surfaces and computes contact kinematics by solving a fixed-size nonlinear system with six equations and six unknowns (Sec.~\ref{sec:math}). This system is solved efficiently (microsecond-scale) using a tailored Newton-based solver that exploits the geometric structure of the contact formulation (Sec.~\ref{sec:solver}). To improve robustness and numerical conditioning, we introduce a scaled surrogate formulation that unifies separating and penetrating configurations of colliding bodies. The robustness and computational efficiency of the approach are evaluated through extensive collision simulations and benchmarking against DCOL (Sec.~\ref{sec:benchmarking}). Finally, IFT-based analytical derivatives are derived from the formulation (Sec.~\ref{sec:analytical_derivatives}) and applied to gradient-based kinematic path planning for a quadrotor and to differentiable contact physics examples, including multibody rigid-body collisions and soft manipulator interactions (Sec.~\ref{sec:applications}). Concluding remarks and discussion are provided in Sec.~\ref{sec:conclusions}.

\section{Related Work}

Given two convex primitives represented by closed convex sets $S_1$ and $S_2$,
DCOL formulates collision detection through uniform scaling about each body’s
reference frame \cite{Tracy2023}. For a scalar scaling factor $\alpha \ge 0$, the scaled sets are
defined as
\[
S_i(\alpha) = \{\, \alpha \bm{x} \mid \bm{x} \in S_i \,\}, \quad i \in \{1,2\}.
\]
\par

Collision detection is posed as finding the smallest uniform scaling factor
$\alpha$ for which the two scaled sets share a common point (Fig. \ref{fig:intro}(a)):
\begin{equation}
\begin{aligned}
\label{eq:DCOL}
\min_{\bm{x},\alpha} \quad & \alpha \\
\text{s.t.} \quad
& \bm{x} \in S_1(\alpha), \\
& \bm{x} \in S_2(\alpha), \\
& \alpha \ge 0 .
\end{aligned}
\end{equation}

The optimal scaling factor $\alpha^*$ serves as an intuitive and continuous collision metric: $\alpha^* > 1$ indicates separation, $\alpha^* = 1$ exact contact, and $\alpha^* < 1$ penetration. The associated solution $\bm{x}^*$ is a witness point lying in the intersection of the two scaled primitives and coincides with the contact points when $\alpha^* = 1$. DCOL efficiently formulates convex primitives commonly encountered in robotics, including polytopes, capsules, and ellipsoids, using set-membership representations based solely on second-order cone and nonnegativity constraints. This structure allows collision detection to be posed as a structured convex optimization problem with conic constraints, which DCOL solves using a custom primal-dual interior-point solver. Subsequent work further interpreted the optimal scaling factor as an optimization-based signed distance function (O-SDF) \cite{LeCleach2023}, with contact normals and related kinematic quantities recovered from the dual variables.

Letting $\bm{z}^*$ denote the primal–dual solution of \eqref{eq:DCOL} and $\bm{\theta}$ a parameter of interest (e.g., generalized coordinates), the KKT conditions define an implicit system $\bm{f}(\bm{z}^*,\bm{\theta})=0$. Assuming $\partial \bm{f}/\partial \bm{z}$ is nonsingular, the IFT yields
\begin{equation}
\label{eq:IFT}
\frac{\partial \bm{z}^*}{\partial \bm{\theta}}
=
- \left( \frac{\partial \bm{f}}{\partial \bm{z}} \right)^{-1}
\frac{\partial \bm{f}}{\partial \bm{\theta}}.
\end{equation}
\par

\section{Contact Geometry and Detection}
\label{sec:math}
Integrating the scaling based optimization of DCOL with implicit surface representation of geometry, we develop iDCOL. We begin by reviewing key properties of implicit surfaces.

\subsection{Derivatives of Transformed Implicit Surfaces}
\begin{figure}[t]
    \centering
    \includegraphics[width=1\linewidth]{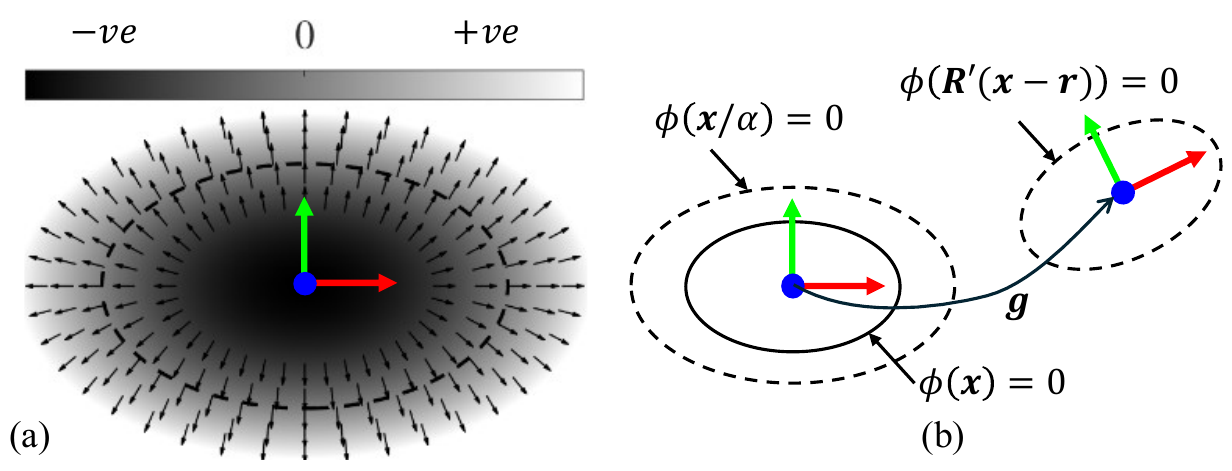}
    \caption{Example of an implicit surface illustrated using a 2D ellipse, $\phi(\bm{x}) = (x_1/a)^2+(x_2/b)^2-1=0$, where $a$ and $b$ are semi-principal axis lengths:
(a) Level sets of the implicit function $\phi(\bm{x})$, with the zero level set
$\phi(\bm{x})=0$ indicated by a dashed curve. Surface normals, given by
$\hat{\bm{n}}(\bm{x})$, are shown at different level sets.
(b) Effects of uniform scaling and rigid-body transformations on implicit
surfaces.}
    \label{fig:implicit_surface}
\end{figure}

An implicit surface is defined as the zero level set of a scalar function
$\phi:\mathbb{R}^3 \to \mathbb{R}$, i.e.,
$\{\bm{x} \mid \phi(\bm{x}) = 0\}$. By convention, $\phi(\bm{x})<0$ denotes the interior of the body and
$\phi(\bm{x})>0$ the exterior. 
The unit normal to the level set of $\phi$ is given by:
\begin{equation}
\label{eq:unit_normal}
\hat{\bm{n}}(\bm{x}) = \frac{\nabla \phi(\bm{x})}{\|\nabla \phi(\bm{x})\| } \, , \quad \text{when} \, \nabla \phi(\bm{x}) \neq \bm{0} \, .
\end{equation}
Fig. \ref{fig:implicit_surface}(a) shows an example of an implicit surface and its normals in 2D. For strictly convex implicit surfaces, the Hessian $\nabla^2 \phi(\bm{x})$, which encodes the local curvature of the level sets, is positive definite on the tangent space of the level set. 

Let $\bm{g}(\bm{q})\in SE(3)$ denote the forward kinematics of a body on a kinematic chain with generalized coordinates $\bm{q}\in\mathbb{R}^{n_{\text{dof}}}$. We have,
\begin{equation}
\bm{g}(\bm{q}) =
\begin{bmatrix}
\bm{R}(\bm{q}) & \bm{r}(\bm{q}) \\
\bm{0}_{1\times3} & 1
\end{bmatrix} 
\end{equation}

Under the rigid-body transformation $\bm g$ and uniform scaling by $\alpha$ (Fig. \ref{fig:implicit_surface}(b)), the implicit surface $\phi:\mathbb{R}^3\to\mathbb{R}$ is expressed in the world frame as
\begin{equation}
\label{eq:transformed_phi}
\phi(\bm{y}) =
\phi\left(\bm{R}^T(\bm{x}-\bm{r})/\alpha\right)
\end{equation}
where, $\bm{y} = \bm{R}^T(\bm{x}-\bm{r})/\alpha$. 

Let $\bm{y}=\bm{y}(u,v)$ denote the transformed coordinates, where $u$ and $v$ represent generic scalar or vector-valued parameters. Using the chain rule of derivatives,

\begin{subequations}
\label{eq:chain_rule_uv}
\begin{align}
\frac{\partial \phi}{\partial u}
&=
\left(\frac{\partial \bm{y}}{\partial u}\right)^T
\nabla \phi(\bm{y}),
\label{eq:chain_rule_first_uv} \\[4pt]
\frac{\partial^2 \phi}{\partial u\,\partial v}
&=
\left(\frac{\partial \bm{y}}{\partial u}\right)^T
\nabla^2 \phi(\bm{y})
\left(\frac{\partial \bm{y}}{\partial v}\right)
+
\left(\frac{\partial^2 \bm{y}}{\partial u\,\partial v}\right)^T
\nabla \phi(\bm{y}).
\label{eq:chain_rule_second_uv}
\end{align}
\end{subequations}
Using, this we derive the first and second order partial derivatives of \eqref{eq:transformed_phi} with respect to $\bm{x}$ and $\alpha$:
\begin{subequations}
\label{eq:phi_partials}
\begin{align}
\phi_{\bm{x}}
&=
\frac{1}{\alpha}\,\bm{R}\,\nabla \phi(\bm{y}),
\label{eq:phi_x} \\[4pt]
\phi_{\alpha}
&=
-\frac{1}{\alpha}\,\bm{y}^T \nabla \phi(\bm{y}),
\label{eq:phi_alpha} \\[6pt]
\phi_{\bm{x}\bm{x}}
&=
\frac{1}{\alpha^2}\,
\bm{R}\,\nabla^2 \phi(\bm{y})\,\bm{R}^T,
\label{eq:phi_xx} \\[6pt]
\phi_{\bm{x}\alpha}
&=
-\frac{1}{\alpha^2}\,
\bm{R}\!\left(
\nabla \phi(\bm{y})
+
\nabla^2 \phi(\bm{y})\,\bm{y}
\right),
\label{eq:phi_xalpha} \\[6pt]
\phi_{\alpha\alpha}
&=
\frac{1}{\alpha^2}\!
\left(
2\,\bm{y}^T \nabla \phi(\bm{y})
+
\bm{y}^T \nabla^2 \phi(\bm{y})\,\bm{y}
\right).
\label{eq:phi_alphaalpha}
\end{align}
\end{subequations}

With analytic surface descriptions and derivatives, trivial scaling and rigid-body transformations, and normals obtained directly from $\nabla \phi$, implicit surfaces provide a natural representation for scaling-based contact kinematics.

\subsection{Contact Optimality Conditions}

Using implicit surface representations, we rewrite \eqref{eq:DCOL} as:

\begin{equation}
\begin{aligned}
\label{eq:iDCOL}
\min_{\bm{x},\alpha} \quad & \alpha \\
\text{s.t.} \quad
& \phi_1(\bm{x}/\alpha) \le 0, \\
& \phi_2(\bm{R}^T(\bm{x}-\bm{r})/\alpha) \le 0 \\
& \alpha \ge 0
\end{aligned} \, .
\end{equation}

Note that in this setting, $\bm{g}$ is the relative rigid-body transformation between body $1$ and body $2$ and $\bm{x}$ is defined in the local frame of body $1$. 

The implicit representation replaces the $N_i$ geometry-dependent conic set-membership constraints of a body in DCOL with a single nonlinear inequality $\phi_i(\cdot)\le 0$, $i=1,2$. When the body-frame origins are not coincident ($\bm{r}\neq \bm{0}$), we have $\alpha^*>0$. Accordingly, the inequality constraint $\alpha\ge0$ is inactive at the solution and does not contribute to the KKT conditions. Under this assumption, the Lagrangian of \eqref{eq:iDCOL} is given by
\begin{equation}
\mathcal{L}(\bm{z})
=
\alpha
+
\lambda_1\,\phi_1(\bm{x}/\alpha)
+
\lambda_2\,\phi_2\!\left(\bm{R}^T(\bm{x}-\bm{r})/\alpha\right),
\end{equation}
where $\bm{z}=[\bm{x}^T,\alpha,\lambda_1,\lambda_2]^T$ and
$\lambda_1,\lambda_2\ge0$ are the Lagrange multipliers (dual) associated with the implicit surface constraints. 
The KKT conditions of \eqref{eq:iDCOL} reduce to the following residual system:
\begin{subequations}
\label{eq:kkt_residual}
\begin{align}
\phi_1 &= 0,
\label{eq:kkt_phi1} \\[2pt]
\phi_2 &= 0,
\label{eq:kkt_phi2} \\[2pt]
\lambda_1\phi_{1\bm{x}} + \lambda_2\phi_{2\bm{x}} &= \bm{0},
\label{eq:kkt_x} \\[2pt]
1 + \lambda_1\phi_{1\alpha} + \lambda_2\phi_{2\alpha} &= 0,
\label{eq:kkt_alpha}
\end{align}
\end{subequations}
where, $\phi_1$ and its partial derivatives  (defined in
\eqref{eq:phi_x}–\eqref{eq:phi_alpha}) are evaluated at $\bm{x}^*/\alpha^*$, and
$\phi_2$ and its derivatives are evaluated at
$\bm{R}^T(\bm{x}^*-\bm{r})/\alpha^*$. Equations \eqref{eq:kkt_phi1}–\eqref{eq:kkt_phi2} correspond to the active implicit surface constraints,  \eqref{eq:kkt_x} enforces stationarity with respect to $\bm{x}$, and \eqref{eq:kkt_alpha} enforces stationarity with respect to $\alpha$. 

Together, these conditions define a system of six scalar equations $\bm{f}_c(\bm{z}^*,\;\bm{q})=\bm{0}$ in six unknowns $\bm{z}^*$, casting iDCOL collision detection as a low-dimensional root-finding problem. The Jacobian of the residual vector $\bm{f}_c$, with respect to $\bm{z}$ is given by:
%
%
\begin{equation}
\label{eq:kkt_jacobian}
\bm{J}_c(\bm{z},\;\bm{q})
=
\begin{bmatrix}
\phi_{1\bm{z}_p}^T
& 0
& 0 \\[2pt]
\phi_{2\bm{z}_p}^T
& 0
& 0 \\[2pt]
\lambda_1 \phi_{1\bm{z}_p\bm{z}_p} + \lambda_2 \phi_{2\bm{z}_p\bm{z}_p}
& \phi_{1\bm{z}_p}
& \phi_{2\bm{z}_p} 
\end{bmatrix}
\end{equation}
where $\bm{z}_p=[\bm{x}^T,\alpha]^T$. The partial derivatives on the RHS are obtained from \eqref{eq:phi_partials}.

Given the analytically computed residual $\bm{f}_c$ and its Jacobian $\bm{J}_c$, the optimal solution can be computed using gradient-based root-finding methods, such as Newton-type algorithms (see Sec. \ref{sec:solver_implementation}). At any solution $\bm{z}^*(\bm q)$ satisfying standard regularity conditions, $\bm{J}_c$ is nonsingular, implying that the IFT can be applied to compute analytical derivatives of $\bm{z}^*$ with respect to $\bm{q}$ via \eqref{eq:IFT} (Sec.~\ref{sec:analytical_derivatives}). Moreover, since strictly convex implicit surfaces admit no edges, corners, or truly flat regions, the associated KKT system admits a generically unique solution.

\subsection{Families of Implicit Convex Primitives}
\label{sec:family}
The iDCOL framework relies on strictly convex implicit surface representations. However, many geometric primitives commonly used in robotics—such as polytopes, cylinders, and cones—are not strictly convex, leading to degeneracies in contact kinematics. 
To restore strict convexity while retaining geometric expressiveness, iDCOL adopts smooth implicit approximations based on
LogSumExp (smooth-max) operator \cite{Nesterov2005} and superquadrics \cite{Barr1981, Lopes2010},
which trade exact geometric fidelity for smoothness and curvature regularity.

LogSumExp provides a smooth approximation of pointwise maximum operators and is
used to blend multiple implicit constraints $c_j(\bm{y})$ — for instance, set-membership constraints defined by convex inequalities as in DCOL primitives \cite{Tracy2023} — into a single strictly convex implicit geometry:
\begin{align}
\phi_{\beta}(\bm{y})&=
\frac{1}{\beta}
\log \sum_{j=1}^m \exp(\beta c_j(\bm y))\\
&=
\max_i c_i
+
\frac{1}{\beta}
\log \sum_{i=1}^m \exp\!\big(\beta (c_i - \max_j c_j)\big) ,\nonumber
\end{align}
where $\beta>0$ controls the sharpness of the approximation. As $\beta \to \infty$, LogSumExp converges to the exact maximum.

\begin{figure}[t]
    \centering
    \includegraphics[width=1\linewidth]{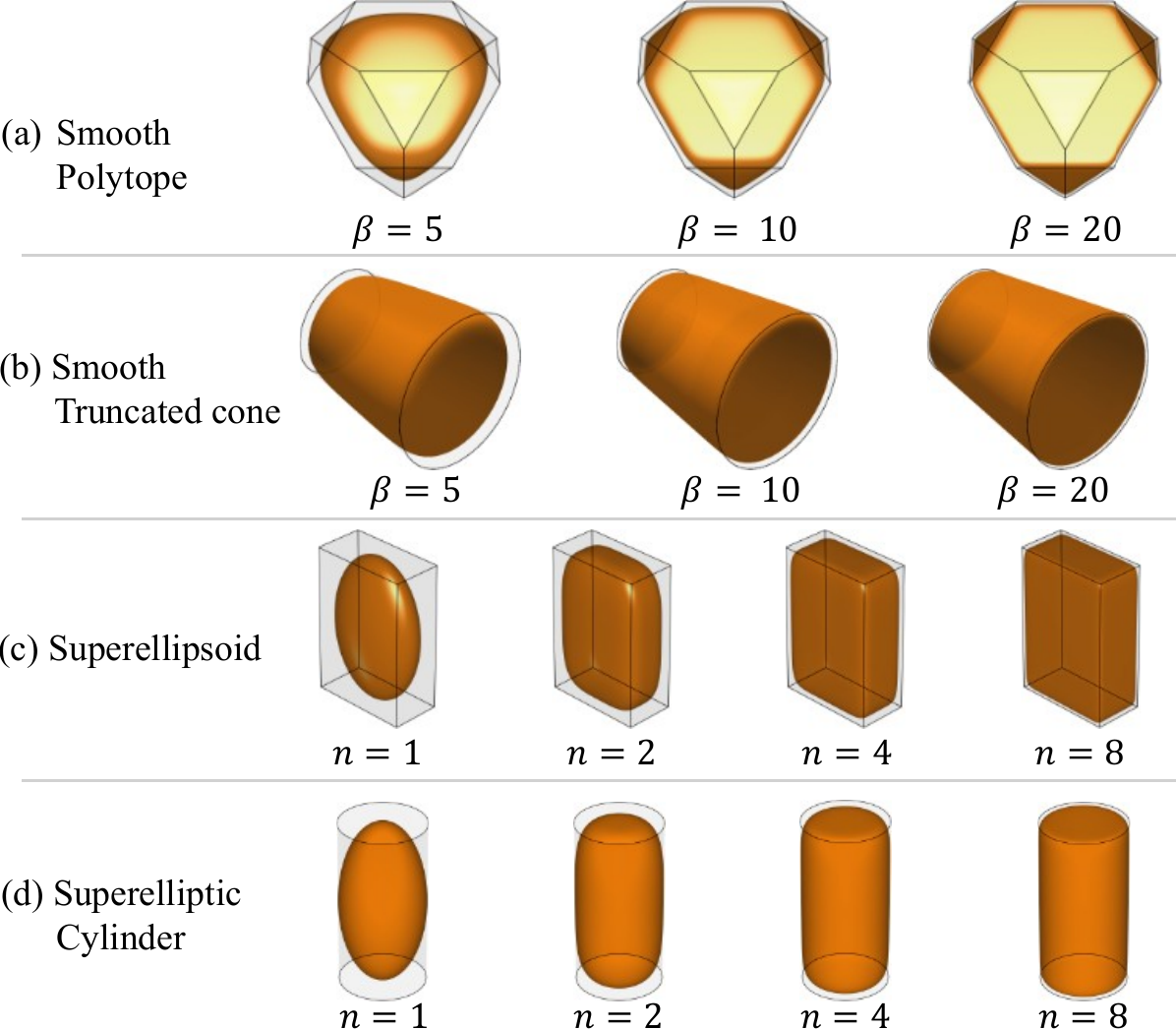}
    \caption{From left to right, increasing $\beta$ or $n$ drives the strictly convex implicit surfaces (orange) to converge to the corresponding exact geometries (black wireframes): (a) smooth polytope, (b) smooth truncated cone, (c) superellipsoid, and (d) superelliptic cylinder.}
    \label{fig:shape_progression}
\end{figure}

The current iDCOL library implements the following families of implicit primitives:

\subsubsection{Smooth Polytope}

A convex polytope can be expressed in half-space form as
\begin{equation}
\mathcal{P} = \{ \bm{y} \in \mathbb{R}^3 \mid \bm{a}_i^T \bm{y} - b_i \le 0,\; i=1,\dots,m \} \, ,
\end{equation}
where $\bm{a}_i \in \mathbb{R}^3$ and $b_i \in \mathbb{R}$ define the $i$-th supporting half-space, and $m$ is the total number of half-space constraints.

While this representation is exact, the resulting geometry is non-strictly convex. To obtain a strictly convex implicit approximation, we replace the pointwise maximum over supporting half-spaces with the LogSumExp operator. 

Let $c_{\mathrm{Poly},i} = \left(\bm{a}_i^T \bm{y} - b_i\right)/L$, where $L > 0$ is a characteristic length for nondimensionalization. The implicit function defining the smooth polytope approximation is given by
\begin{equation}
\phi_{\mathrm{Poly}}(\bm{y})
=
\operatorname{smax}_{\beta}(\bm{c}_\mathrm{Poly}) \, .
\end{equation}

This construction provides a smooth $C^\infty$ approximation of arbitrary convex polytopes, including boxes, prisms, pyramids, and general convex hulls defined by planar facets.

\subsubsection{Smooth Truncated Cone}

A truncated cone with base radius $R_b$, top radius $R_t$, and axial extents $[-a,b]$ can be described exactly as the intersection of three implicit inequalities,
\begin{equation}
\frac{y_2^2 + y_3^2}{R^2(y_1)} - 1 \le 0,
\qquad
-\frac{y_1}{a} - 1 \le 0,
\qquad
\frac{y_1}{b} - 1 \le 0,
\end{equation}
where the three inequalities correspond to
$\phi_{\mathrm{side}}(\bm{y})$,
$\phi_{\mathrm{bot}}(\bm{y})$,
and $\phi_{\mathrm{top}}(\bm{y})$, respectively, and the radius varies linearly along the axis as
\begin{equation}
R(y_1) = R_b + (R_t - R_b)\frac{y_1 + a}{a + b}.
\end{equation}

To obtain a smooth implicit approximation compatible with iDCOL, these components are combined using the LogSumExp operator with $\bm{c}_\mathrm{TC} = [\phi_{\mathrm{side}},\phi_{\mathrm{bot}},\phi_{\mathrm{top}}]^T$,
\begin{equation}
\phi_{\mathrm{TC}}(\bm{y})
=
\operatorname{smax}_{\beta}(
\bm{c}_\mathrm{TC}) \, .
\end{equation}

This construction provides a strictly convex approximation of cones (with $R_t$ regularized by a small positive value) and frusta, enabling the modeling of tapered robotic links, nozzles, and axisymmetric components with continuously varying cross-sections.

\subsubsection{Superellipsoid}

Superellipsoids form a family of implicit primitives that can represent a wide range of rounded geometries \cite{Lopes2010, Gonalves2017}. The implicit surface is defined as
\begin{equation}
\phi_{\mathrm{SE}}(\bm{y})
=
\left(\left( \frac{y_1}{a} \right)^{2n}
+
\left( \frac{y_2}{b} \right)^{2n}
+
\left( \frac{y_3}{c} \right)^{2n}\right)^{\frac{1}{2n}}
- 1 \, ,
\end{equation}
where $a,b,c>0$ are scaling parameters along the principal axes and $n \in \mathbb{N}$ controls the shape exponent.

For $n=1$, the superellipsoid reduces to an ellipsoid. As $n$ increases, the shape becomes progressively box-like while remaining strictly convex, and in the limit $n \to \infty$ it converges point-wise to an axis-aligned box. This family smoothly interpolates between ellipsoids and boxes, allowing exact modeling of spheres and ellipsoids, and strictly convex approximations of box-like robot geometries.

\subsubsection{Superelliptic Cylinder}

To represent cylindrical geometries with smooth boundaries, we use a superelliptic cylinder defined by the implicit function
\begin{equation}
\phi_{\mathrm{SEC}}(\bm{y})
=
\left(\left( \frac{y_2^2 + y_3^2}{R^2} \right)^{n}
+
\left( \frac{y_1}{h} \right)^{2n}\right)^{\frac{1}{2n}}
- 1 \, ,
\end{equation}
where $R>0$ is the radial scale, $h>0$ is the axial half-length, and $n \in \mathbb{N}$ controls the shape exponent.

For $n=1$, this formulation reduces to a smooth ellipsoidal cylinder. As $n$ increases, the radial and axial profiles become progressively flatter while remaining smooth, and in the limit $n \to \infty$ the shape converges to a finite circular cylinder. This family enables strictly convex approximations of cylinders, capped cylinders, and elongated convex bodies.

The progression of the proposed implicit convex primitives as the sharpness parameters $\beta$ or $n$ increase is illustrated in Fig. \ref{fig:shape_progression}. As these parameters grow, the strictly convex surfaces converge to their corresponding exact geometric counterparts while remaining differentiable. The analytical expressions for $\nabla \phi$ and $\nabla^2 \phi$, for all primitives introduced in this section, are provided in the appendix \ref{app:shapes}.

\section{Numerical Solution Strategy}
\label{sec:solver}
Although the contact problem is expressed as a system of nonlinear equations via \eqref{eq:kkt_residual} and \eqref{eq:kkt_jacobian}, direct application of a Newton-type solver can still be unreliable. This section summarizes the key numerical strategies used to ensure robustness and efficiency.

\subsection{Scaling Reparameterization and Bounds}
\label{sec:re_para_bounds}
We reparameterize the scaling variable as $\alpha=e^{s}$ with $s\in\mathbb{R}$, which enforces $\alpha>0$ and improves conditioning. The decision variable becomes $\bm{z}=[\bm{x}^T,s,\lambda_1,\lambda_2]^T$. The form of the KKT residual \eqref{eq:kkt_residual} remains unchanged while the Jacobian, corresponding to the scaling variable, is modified via the chain rule: $(\cdot)_s = (\cdot)_\alpha\alpha$.  

Geometric bounds on $\alpha^*$ are derived using concentric inner and outer
bounding spheres for each body (Fig.~\ref{fig:geometry_insights}(a)). Let
$r_{i,\mathrm{in}}$ and $r_{i,\mathrm{out}}$ denote their radii (precomputed once per body). For a given
relative translation $\bm{r}$,
\begin{equation}
\alpha_{\min} = \frac{\|\bm{r}\|}{r_{1,\mathrm{out}} + r_{2,\mathrm{out}}},
\qquad
\alpha_{\max} = \frac{\|\bm{r}\|}{r_{1,\mathrm{in}} + r_{2,\mathrm{in}}} \, .
\end{equation}

These bounds define a trust region for $s=\log\alpha$ and significantly improve
conditioning, particularly for nearly spherical bodies.


\subsection{Surrogate Formulation}
\label{sec:surrogate}
\begin{figure}[t]
    \centering
    \includegraphics[width=1\linewidth]{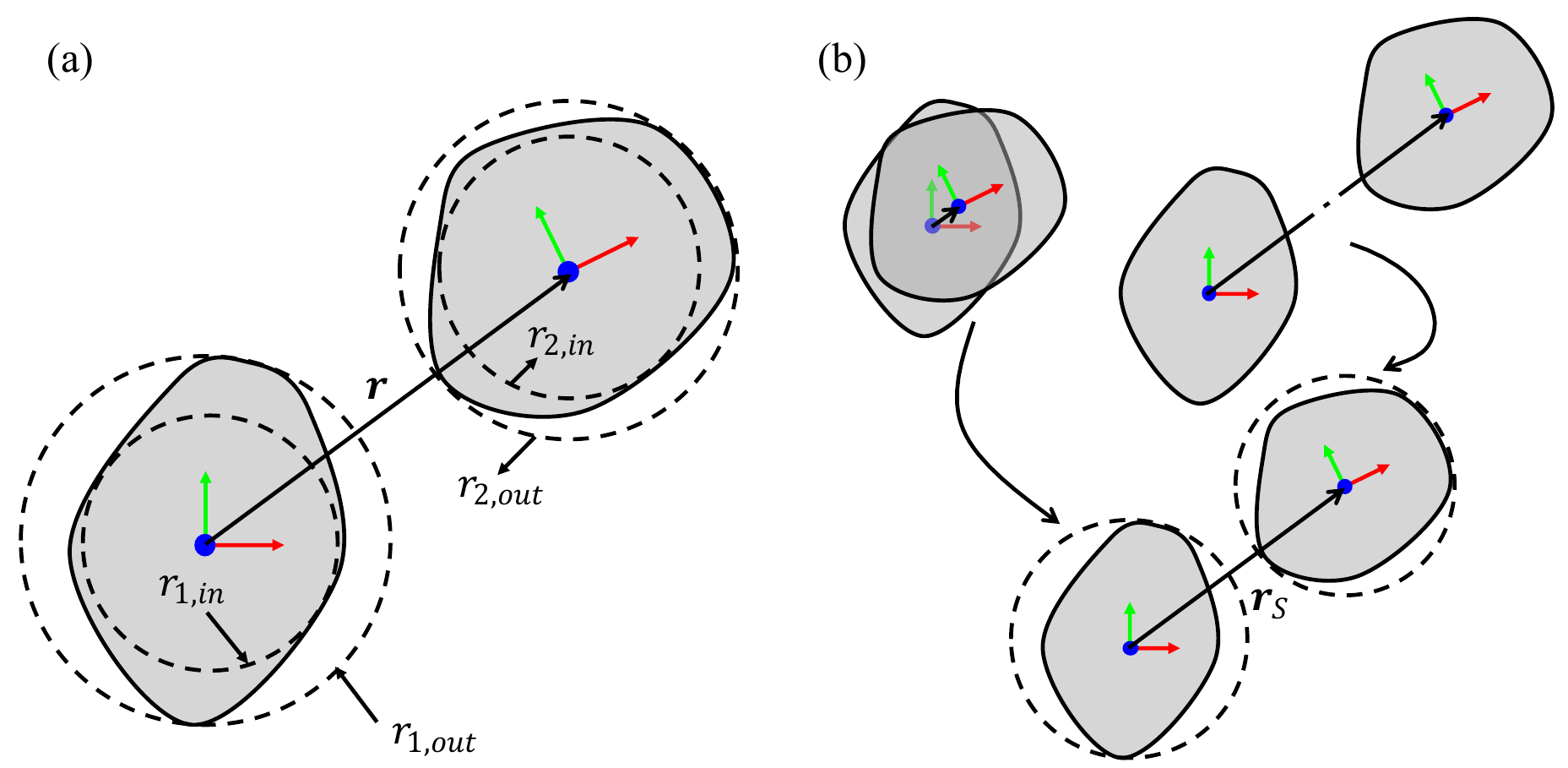}
    \caption{Geometrical bounds and surrogate formulation: (a) A contact body pair with inner and outer bounding spheres centered at the body frame origin. (b) A surrogate problem is introduced to handle configurations that are either near-coincident or widely separated. The problem rescales the relative translation such that the bounding spheres are non-penetrating.}
    \label{fig:geometry_insights}
\end{figure}

To handle configurations in which $\alpha^*$ becomes very small (nearly coincident) or very large (widely separated), we introduce a surrogate problem that rescales the relative translation, as illustrated Fig.~\ref{fig:geometry_insights}(b):
\begin{equation}
    \bm{r}_S = \bm{r}/(\alpha_{\min}/f_S)=f_S(r_{1,\mathrm{out}} + r_{2,\mathrm{out}})\hat{\bm{r}}
\end{equation}
where $f_S\ge1$ is a separating factor and the subscript $(\cdot)_S$ denotes quantities associated with the surrogate problem.

The rationale for this surrogate formulation is as follows. When the bodies are close, scaling down both bodies by $\alpha_{\min}$ guarantees a non-penetrating configuration (tangent outer spheres when $f_S = 1$) and is equivalent to scaling up the relative translation by $1/\alpha_{\min}$. Once solved, the solution of the original problem is recovered as $\alpha^* = (\alpha_{\min}/f_S)\alpha_S^*$, $\bm{x}^* = (\alpha_{\min}/f_S)\bm{x}_S^*$, and $\lambda_i^* = (\alpha_{\min}/f_S)\lambda_{iS}^*$. A similar argument applies when the bodies are widely separated. For the surrogate problem, the scaling bounds become:
\begin{equation}
\label{eq:surrogate_bounds}
\alpha_{\min,S} = f_S \, , \qquad \alpha_{\max,S} =
                f_S\frac{r_{1,\mathrm{out}} + r_{2,\mathrm{out}}}{r_{1,\mathrm{in}} + r_{2,\mathrm{in}}},
\end{equation}
which are independent of the relative pose, with the lower bound also being independent of the geometry.

\textbf{Remark}: The surrogate problem treats separating and penetrating configurations identically and places rigid bodies in a well-conditioned relative position for collision detection.

\subsection{Numerical Solver}

\label{sec:solver_implementation}
We develop a custom safeguarded Newton–type method for solving the surrogate KKT
system \eqref{eq:kkt_residual} using analytically computed Jacobians
\eqref{eq:kkt_jacobian}. At each iteration, a Newton step is computed by solving
the linear system and globalized using a backtracking Armijo line search on the
merit function $m(\bm z)=\tfrac{1}{2}\|\bm{f}_c(\bm z)\|^2$. Damped least-squares updates are employed when necessary to maintain descent and improve numerical
robustness. The overall solver structure is summarized in Algorithm~\ref{alg:idcol_solver}. 

\begin{algorithm}[t]
\caption{Safeguarded Newton Solver for iDCOL. $\mathcal{P}$ is the problem data, $N_{\mathrm{a}}$ is the number of restart attempts, and $k_{\max}$ is the maximum number of Newton iterations.}
\label{alg:idcol_solver}
\begin{algorithmic}[1]
\Require Initial guess $\bm{z}_0=(\bm{x}_{S,0},s_{S,0},\lambda_{1S,0},\lambda_{2S,0})$, $\mathcal{P}$
\Ensure Approximate solution $\bm{z}^*$
\For{$a = 1,2,\dots,N_{\mathrm{a}}$}
  \State $\bm{z}\gets \bm{z}_0$
  \For{$k = 0,1,\dots,k_{\max}$}
    \State Evaluate $\bm{f}_c(\bm{z})$ \eqref{eq:kkt_residual} and $\bm{J}_c(\bm{z})$ \eqref{eq:kkt_jacobian}
    \If{$\|\bm{f}_c(\bm{z})\| < \mathrm{tol}$}
      \State \Return $\bm{z}$
    \EndIf
    \State Solve $\bm{J}_c(\bm{z})\Delta\bm{z}=-\bm{f}_c(\bm{z})$
    \If{Newton step is ill-conditioned}
      \State Compute LM fallback step
    \EndIf
    \State Apply step-size limits
    \State Backtracking line search on $m(\bm{z})=\frac{1}{2}\|\bm{f}_c(\bm{z})\|^2$
    \If{step accepted}
      \State $\bm{z}\gets \bm{z}+\Delta\bm{z}$
    \Else
      \State Use damped LM-based fallback update
    \EndIf
  \EndFor
  \State Perturb $s_{S,0}$ and retry
\EndFor
\State \Return best iterate encountered
\end{algorithmic}
\end{algorithm}

\paragraph{Initialization.}
When a previous solution is available (e.g., in dynamic simulations), it can be used to warm-start the solver. Otherwise, the surrogate geometry provides a natural initialization: $\bm{x}_{S,0}$ is initialized along $\hat{\bm{r}}$ on the outer bounding sphere, and the scaling variable is initialized as the geometric mean of its bounds \eqref{eq:surrogate_bounds}, $\alpha_{S,0} = \sqrt{\alpha_{\min,S}\alpha_{\max,S}}$. The initial Lagrange multipliers $\lambda_{1S,0}$ and $\lambda_{2S,0}$ are then obtained by substituting $\bm{x}_{S,0}$ and $\alpha_{S,0}$ into the stationarity conditions \eqref{eq:kkt_x} and \eqref{eq:kkt_alpha}. The resulting cold-start initialization is
\begin{align}
\label{eq:guess}
        \bm{x}_{S,0}
&= r_{1,\mathrm{out}}\hat{\bm{r}},
\qquad
s_{S,0}
= \tfrac{1}{2}\log(\alpha_{\max,S}\alpha_{\min,S}),
\\
\lambda_{1S,0}
&= \frac{\alpha_{S,0}}{\bm{r}_S^T \phi_{1\bm{x}_S}},
\qquad
\lambda_{2S,0}
= \frac{-\alpha_{S,0}}{\bm{r}_S^T \phi_{2\bm{x}_S}} . \nonumber
\end{align}

\textbf{Remark}: For sphere–sphere contacts, \eqref{eq:guess} coincides with the exact solution.

\paragraph{Continuation strategy.}
\label{sec:continuation}
In challenging cases involving sharp geometries, an optional continuation strategy can be employed to further improve robustness. If convergence fails at the target shape parameters, the problem is first solved for a smoother instance (smaller $\beta$ or $n$), and the resulting solution is used to initialize progressively sharper problems.

\begin{figure}[t]
    \centering
    \includegraphics[width=1\linewidth]{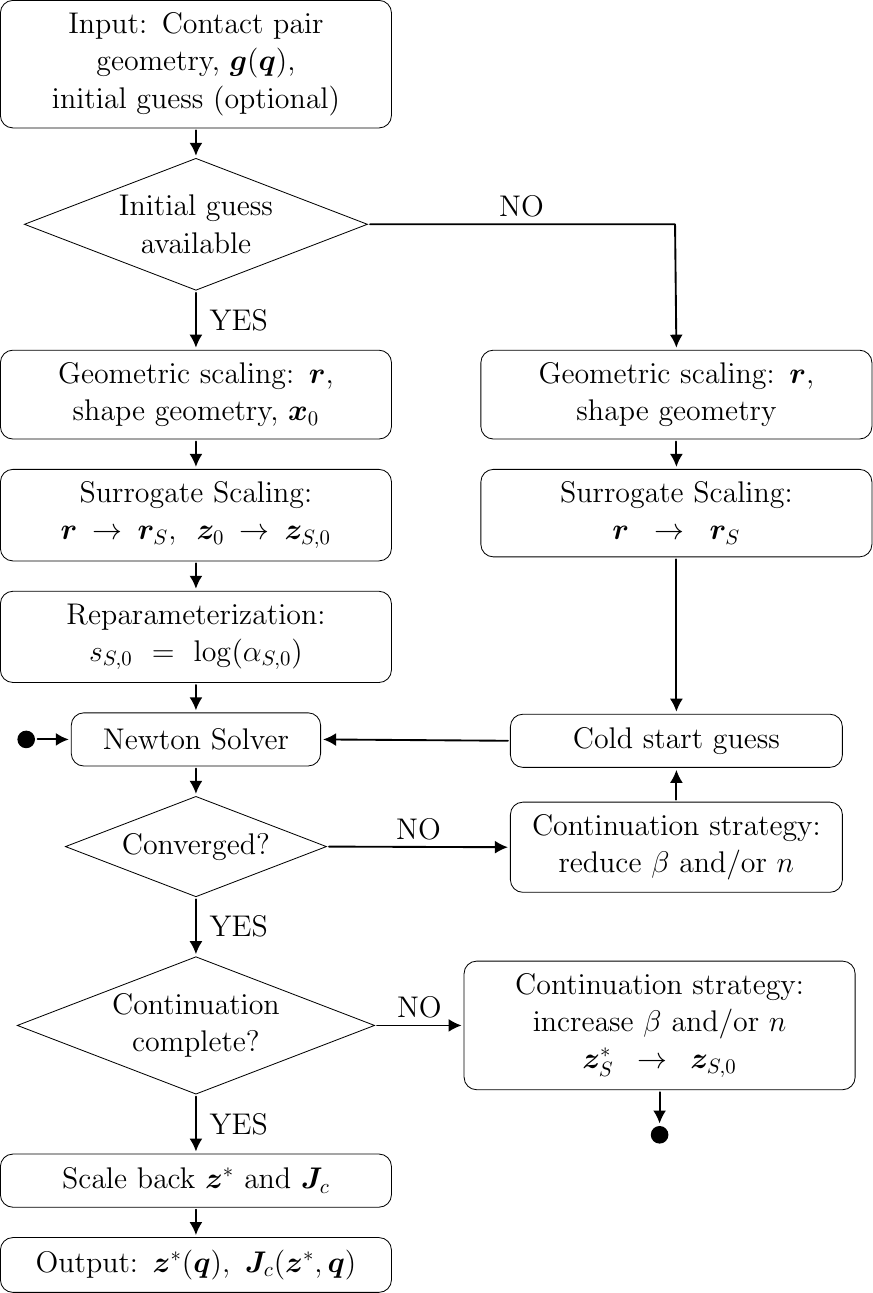}
    \caption{Schematic of the iDCOL numerical solver pipeline. Numerical robustness is improved through geometric scaling, surrogate scaling (Sec.~\ref{sec:surrogate}), and reparameterization of the scaling variable $\alpha$ (Sec.~\ref{sec:re_para_bounds}). Geometric scaling uniformly rescales the contact geometry by a fixed length scale (e.g., $\max(r_{1,\mathrm{out}},, r_{2,\mathrm{out}})$). The Newton Solver block corresponds to Algorithm~\ref{alg:idcol_solver}. The continuation strategy is triggered if the Newton solver fails to converge or if continuation parameters (when present) have not yet reached their target values.}
    \label{fig:solver_flowchart}
\end{figure}

The overall iDCOL solver architecture is illustrated in Fig.~\ref{fig:solver_flowchart}. The solver takes as input the geometric information of contact pairs, the relative pose $\bm{g}$, and an optional initial guess, and returns the optimal solution $\bm{z}^*$ together with the corresponding Jacobian $\bm{J}_c$ required for IFT application. The complete iDCOL framework, including geometry evaluation and solver components, is implemented in C++~\cite{iDCOLcode}.

\section{Simulations and Benchmarking}
\label{sec:benchmarking}
For all simulations, the contact solver is evaluated using a deterministic sweep that asymptotically explores the relative pose space $SE(3)$. Relative position and orientation evolve at mutually incommensurate frequencies, producing an ergodic, non-repeating trajectory with $10^6$ evaluated poses per scenario, providing dense coverage of contact configurations. All computations are performed on a machine equipped with a 13th Gen Intel(R) Core(TM) i9-13900HX processor (2.20 GHz) and 64 GB of RAM.

\subsection{iDCOL Robustness and Runtime}

\begin{table}[t]
\centering
\caption{Average runtime ($\mu$s) for iDCOL across shape pairs.
Top value: no warm start.
\textbf{Bottom value}: with warm start.}
\label{tab:runtime_all}
\renewcommand{\arraystretch}{1.2}
\begin{tabular}{c|cccc}
\hline
Body 1 $\backslash$ Body 2 
& Poly 
& TC
& SE
& SEC \\
\hline
Poly 
& \begin{tabular}[c]{@{}c@{}}4.73\\ \textbf{2.13}\end{tabular}
& \begin{tabular}[c]{@{}c@{}}4.26\\ \textbf{1.94}\end{tabular}
& \begin{tabular}[c]{@{}c@{}}4.80\\ \textbf{1.96}\end{tabular}
& \begin{tabular}[c]{@{}c@{}}4.21\\ \textbf{1.85}\end{tabular} \\

TC
& -- 
& \begin{tabular}[c]{@{}c@{}}4.12\\ \textbf{1.88}\end{tabular}
& \begin{tabular}[c]{@{}c@{}}5.05\\ \textbf{1.91}\end{tabular}
& \begin{tabular}[c]{@{}c@{}}3.81\\ \textbf{1.74}\end{tabular} \\

SE
& -- 
& -- 
& \begin{tabular}[c]{@{}c@{}}19.39\\ \textbf{1.84}\end{tabular}
& \begin{tabular}[c]{@{}c@{}}17.64\\ \textbf{1.70}\end{tabular} \\

SEC 
& -- 
& -- 
& -- 
& \begin{tabular}[c]{@{}c@{}}5.78\\ \textbf{1.64}\end{tabular} \\
\hline
\end{tabular}
\end{table}

Table~\ref{tab:runtime_all} reports iDCOL runtimes across all shape pairs, with and without warm start, using the primitives in Fig.~\ref{fig:shape_progression} ($\beta=20,\; n=8$). Without warm start, the solver converges reliably in all cases, with average runtimes reflecting the geometric complexity and nonlinearity of the underlying shape representations. Shape pairs involving the superelliptic family (SE and SEC) exhibit higher average cold-start runtimes due to the increased nonlinearity induced by $n$. In these cases, the continuation strategy described in Sec.~\ref{sec:solver_implementation} is activated to ensure convergence, which contributes to the additional computational cost. Note that, despite the higher averages, the median cold-start runtimes for SE-based pairs are comparable to those of other shape combinations, indicating that the overhead arises from a small number of difficult configurations rather than typical behavior. With a warm start, all shape pairs exhibit a consistent reduction in runtime, achieving average solve times near $2~\mu$s with low iteration counts.

\begin{figure}[t]
    \centering
    \includegraphics[width=0.9\linewidth]{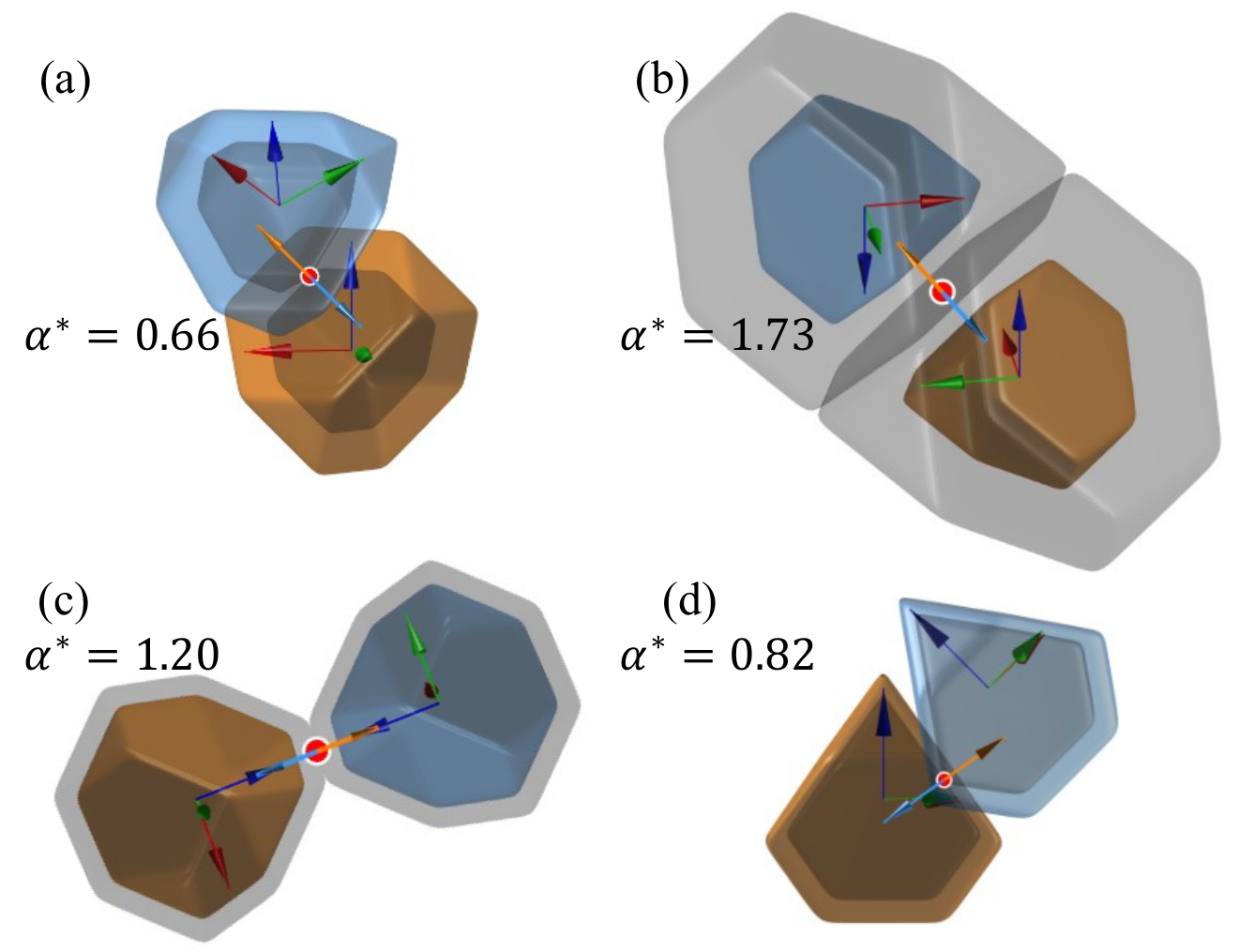}
    \caption{Examples of polytope--polytope contact configurations.
(a) Arbitrary contact configuration.
(b) Face--face contact.
(c) Edge--edge contact with parallel edges.
(d) Face--edge contact with parallel features.
The optimal contact point (red spot, $\bm{x}^*$) and associated normals are shown in each case. Cases (a) and (d) correspond to penetrating configurations ($\alpha^* < 1$), while (b) and (c) correspond to separated configurations ($\alpha^* > 1$).}

    \label{fig:examples}
\end{figure}

Figure~\ref{fig:examples} illustrates representative polytope--polytope contact configurations, including face--face, edge--edge, and face--edge contacts. Both penetrating and separated configurations are shown, demonstrating the solver’s ability to robustly handle diverse contact modes within a unified formulation.

\subsection{Comparison with DCOL}

We compare iDCOL with DCOL using the public Julia implementation \cite{DCOLcode}, evaluating numerical behavior and runtime on three representative cases: polytope–polytope, polytope–ellipsoid (Ellip., $\mathrm{SE}$ with $n=1$), and ellipsoid–ellipsoid, noting that DCOL operates on exact polyhedral geometry. Results are summarized in Table~\ref{tab:dcol_idcol_runtime}.

\begin{table}[t]
\centering
\caption{Average runtime per contact query ($\mu$s).
For iDCOL, the top value is without warm start, and the \textbf{bottom value} is with warm start.}
\label{tab:dcol_idcol_runtime}
\renewcommand{\arraystretch}{1.2}
\begin{tabular}{c|ccc}
\hline
Method & Poly--Poly & Poly--Ellip. & Ellip.--Ellip. \\
\hline
DCOL
& 2.29
& 4.06
& 3.57 \\

iDCOL
& \begin{tabular}[c]{@{}c@{}}4.73\\ \textbf{2.13}\end{tabular}
& \begin{tabular}[c]{@{}c@{}}3.98\\ \textbf{1.85}\end{tabular}
& \begin{tabular}[c]{@{}c@{}}2.58\\ \textbf{1.33}\end{tabular} \\
\hline
\end{tabular}
\end{table}

Without warm start, iDCOL is about $2.1\times$ slower than DCOL for polytope–polytope contacts, while achieving comparable performance for polytope–ellipsoid and faster performance for ellipsoid–ellipsoid pairs. With a warm start, iDCOL reduces runtime by roughly $2\times$ across all cases, becoming comparable to DCOL for polytope–polytope and consistently faster for the other two cases.
Overall, the results indicate a clear geometry-dependent trend: DCOL, due to its linear non-negativity constraints, is faster for polytope-like geometries. In contrast, smoother shape pairs increasingly favor iDCOL in this benchmark.

For ellipsoid–ellipsoid contacts, both methods converge to identical contact points $\bm{x}^*$ and scaling factors $\alpha^*$ (up to numerical tolerance), validating the iDCOL formulation. For the other two cases, the solutions differ, but iDCOL yields smoother configuration dependence (regulated by $\beta$ or $n$), which is advantageous for differentiable simulation and optimization.

\section{Analytical Derivatives of Contact Kinematics}
\label{sec:analytical_derivatives}

We compute the analytical derivatives of $\bm{z}^*$ with respect
to $\bm{q}$ by applying IFT \eqref{eq:IFT} on \eqref{eq:kkt_residual}. This requires evaluating $\partial \bm{f}_c / \partial \bm{q}$. Once
$\partial \bm{z}^* / \partial \bm{q}$ is available, derivatives of any contact kinematic quantity
$\bm{h}$ (e.g., contact point, signed distance, or contact normal) follow directly by
differentiation of $\bm{h}(\bm{z}^*(\bm{q}),\bm{q})$.

\subsection{Derivative of the Optimal Solution With Respect to the Configuration}

For $\bm{y} = \bm{R}^T(\bm{x}-\bm{r})/\alpha$, we have:
\begin{equation}
\label{eq:y_partials_q}
\bm{y}_{\bm{q}}
=
\left[
\tilde{\bm{y}}
\;\;
-\frac{1}{\alpha}\bm{I}_3
\right]\bm{J},
\;\;
\bm{y}_{\bm{x}\bm{q}}
=
\frac{1}{\alpha}\,\frac{\partial \bm{R}^T}{\partial \bm{q}},
\;\;
\bm{y}_{\alpha\bm{q}} = -\frac{1}{\alpha}\bm{y}_{\bm{q}}
\end{equation}
where $\tilde{(\cdot)}:\mathbb{R}^3 \rightarrow \mathfrak{so}(3)$ denotes the skew-symmetric operator, and the geometric Jacobian $\bm{J}(\bm q) \in \mathbb{R}^{6 \times n_{\text{dof}}}$ maps infinitesimal variations in $\bm{q}$ to variations of $\bm{g}(\bm{q})$ in the tangent space of $SE(3)$, i.e., $(\bm{g}^{-1}\delta \bm{g})^\vee =\bm{J}\delta \bm{q}$, with $(\cdot)^\vee : \mathfrak{se}(3) \rightarrow \mathbb{R}^6$ \cite{ModernRobotics}. 

Applying \eqref{eq:chain_rule_uv} on $\phi(\bm{y})$ we get:
\begin{subequations}
\label{eq:phi_partialsq}
\begin{align}
\phi_{\bm{q}}
&=
\nabla \phi(\bm{y})^T
\left[
\tilde{\bm{y}}
\;\;
-\frac{1}{\alpha}\bm{I}_3
\right]\bm{J},
\label{eq:phi_q}
\\
\phi_{\bm{x}\bm{q}}
&=\frac{1}{\alpha}\bm{R}
\begin{bmatrix}
\nabla^2\phi(\bm{y})\tilde{\bm{y}}-\widetilde{\nabla\phi(\bm{y})}
&
-\frac{1}{\alpha}\nabla^2\phi(\bm{y})
\end{bmatrix}
\bm{J},
\label{eq:phi_xq}
\\
\phi_{\alpha\bm{q}}
&=
-\frac{1}{\alpha}
\left(
\nabla \phi(\bm{y})
+
\nabla^2 \phi(\bm{y})\,\bm{y}
\right)^T
\left[
\tilde{\bm{y}}
\;\;
-\frac{1}{\alpha}\bm{I}_3
\right]\bm{J}.
\label{eq:phi_alphaq}
\end{align}
\end{subequations}

Using \eqref{eq:phi_partialsq}, we compute the partial derivative of \eqref{eq:kkt_residual} with respect to $\bm{q}$.
While $\phi_1(\bm{x}/\alpha)$ is independent of $\bm{q}$, $\phi_2(\bm{R}^T(\bm{x}-\bm{r})/\alpha)$ depends explicitly on $\bm{q}$. We have,

\begin{equation}
\bm{f}_{c,\bm{q}}
=
\begin{bmatrix}
\bm{0}
\\[2pt]
\phi_{2\bm{q}}
\\[2pt]
\lambda_2\,\phi_{2\bm{x}\bm{q}}
\\[2pt]
\lambda_2\,\phi_{2\alpha\bm{q}}
\end{bmatrix} = \underbrace{
\begin{bmatrix}
\bm{0}
\\[2pt]
\bm{a}^T
\\[2pt]
\lambda_2\,\bm{B}
\\[2pt]
\lambda_2\,\bm{c}^T
\end{bmatrix}}_{\bm{G}_{\mathrm{c}}(\bm{z}^*, \bm{q})}
\,\bm{J},
\label{eq:df_dq}
\end{equation}
where $\bm{a}$, $\bm{B}$, and $\bm{c}$ are obtained from \eqref{eq:phi_partialsq}. 

Applying IFT \eqref{eq:IFT} on $\bm{f}_{c}(\bm{z}^*, \bm{q}) = \bm{0}$ we get:
\begin{align}
\label{eq:dz_dq}
\frac{\partial \bm{z}^*}{\partial \bm{q}}
&= -\bm{J}_c^{-1}\,\bm{f}_{c,\bm{q}}= -\bm{J}_c^{-1}\,\bm{G}_c\,\bm{J} \\
&= \bm{T}_c\,\bm{J}\nonumber \, .
\end{align}

In practice, $\bm{T}_c$ can be evaluated by solving the linear system
\begin{equation}
\bm{J}_c\bm{T}_c = -\bm{G}_c,
\end{equation}
reusing the factorization of $\bm{J}_c$ computed during the Newton solve. As a result, analytical derivatives \eqref{eq:dz_dq} incur negligible overhead relative to collision detection itself, whose runtime is already on the order of microseconds (Table \ref{tab:runtime_all}).

\subsection{Contact Kinematic Quantities}
\begin{figure}[t]
    \centering
\includegraphics[width=0.6\linewidth]{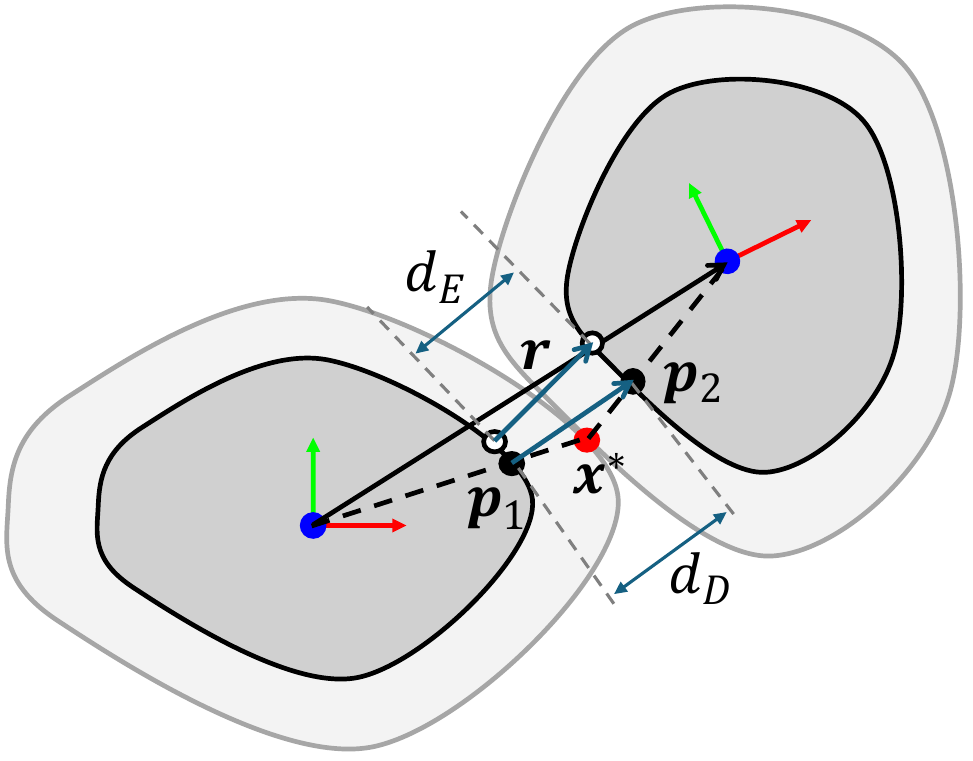}
    \caption{Witness points and distance induced by the iDCOL formulation. The witness points (black filled circles) $\bm{p}_1$ and $\bm{p}_2$ are obtained by scaling back the intersection point $\bm{x}^*$ of the uniformly scaled bodies to the original bodies. The Euclidean closest-point distance $d_E$ and the corresponding closest points (unfilled circles) are shown for comparison.}
    \label{fig:dcol_gap}
\end{figure}

Figure~\ref{fig:dcol_gap} illustrates the contact geometry induced by the iDCOL formulation.
Given the optimal scaling factor $\alpha^*$ and the intersection point $\bm{x}^*$ of the uniformly scaled bodies,
the witness points $\bm{p}_1$ and $\bm{p}_2$ are obtained by scaling back $\bm{x}^*$ to the original bodies. Specifically, expressed in the frame of body~1, the witness points are given by
\begin{equation}
\label{eq:witness}
\bm{p}_1 = \frac{1}{\alpha^*}\,\bm{x}^*,
\qquad
\bm{p}_2 = \left(1-\frac{1}{\alpha^*}\right)\,\bm{r}+\frac{1}{\alpha^*}\,\bm{x}^*
\end{equation}

Using these points, we define the iDCOL gap function as follows:
\begin{equation}
\label{eq:gap}
d_D = \left(1-\frac{1}{\alpha^*}\right)\|\bm{r}\|
\end{equation}

The quantity $d_D$ defines a separation ($d_D>0$) or penetration ($d_D<0$) measure induced by the iDCOL uniform scaling construction.
While the witness points and the associated distance are well-defined and smooth,
$d_D$ does not, in general, coincide with the Euclidean closest-point distance $d_E$.
Nevertheless, $d_D$ provides a differentiable penetration measure that can be directly used in penalty-based contact constitutive models \cite{Flores2022}. At contact, $\alpha^* = 1$, and by~\eqref{eq:gap} we have $d_D = d_E = 0$.

Another contact kinematic quantity of interest is the unit contact normal, defined by \eqref{eq:unit_normal}, which is evaluated at the intersection point $\bm{x}^*$ of the uniformly scaled bodies. Analytical derivatives of the witness points \eqref{eq:witness}, the gap function \eqref{eq:gap}, and the unit normal \eqref{eq:unit_normal} can be obtained via the chain rule using \eqref{eq:dz_dq}. Their explicit formulas are provided in the appendix \ref{app:contact}.

\section{Applications}

\label{sec:applications}
We exploit the differentiability of iDCOL in a kinematic path planning task and two differentiable contact physics examples. All simulations are implemented in MATLAB with efficient compiled collision routines (MEX). Animations of all examples are provided in the supplementary material.

\subsection{Quadrotor Path Planning}
\begin{figure}[t]
    \centering
    \includegraphics[width=\linewidth]{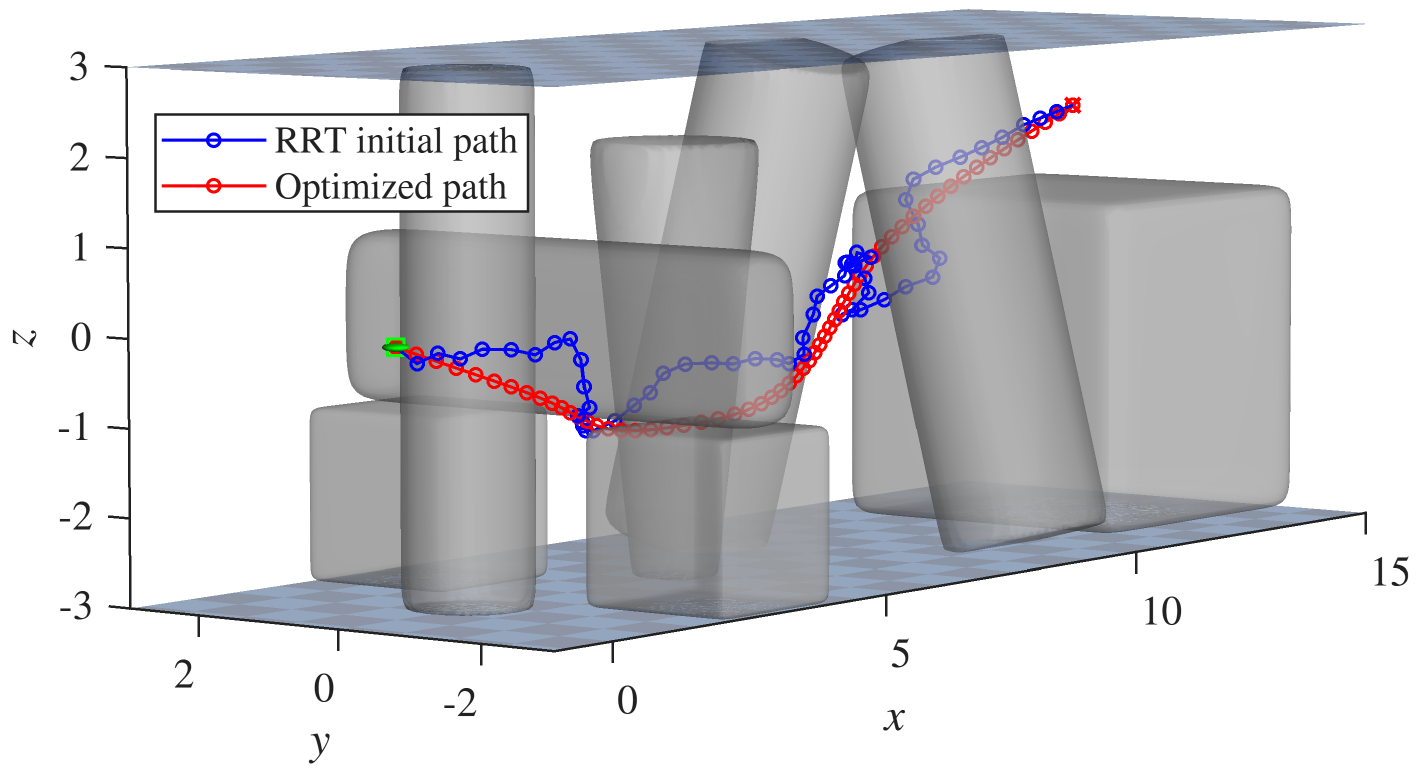}
    \caption{
    The initial collision-free path obtained via RRT (blue) is refined using gradient-based optimization with iDCOL, yielding a smooth optimized trajectory (red). Transparent primitives depict the environment geometry, while the checker planes indicate workspace boundaries.
    }
    \label{fig:opt_path}
\end{figure}

We consider a translation-only kinematic trajectory optimization problem for a quadrotor navigating a cluttered environment (Fig.~\ref{fig:opt_path}). The orientation is fixed, reducing the generalized coordinates to the translational position $\bm{q}\equiv\bm{p}\in\mathbb{R}^3$, which is optimized along the path. The quadrotor is modeled as a single ellipsoidal collision primitive, and the environment comprises eight static convex obstacles.

A collision-free initial path is obtained using the rapidly-exploring random tree (RRT)~\cite{lavalle2001randomized} and refined via a nonlinear program solved with \texttt{fmincon} using the interior-point method. The objective penalizes path length and rewards smoothness, while collision avoidance is enforced via differentiable inequality constraints derived from iDCOL: $\alpha^*(\bm{p})\ge1+\epsilon$. Analytical gradients of the objective and constraints are supplied, with constraint gradients computed using \eqref{eq:dz_dq}. The full optimization requires approximately $10$~s on average, depending on the RRT initialization, with collision queries accounting for roughly $17\%$ of the total computation time. Without analytical gradients, the optimization time increases by nearly two orders of magnitude.

\subsection{Differentiable Contact Physics}
\label{sec:contact_physics}

\begin{figure*}[t]
    \centering
    \includegraphics[width=\textwidth]{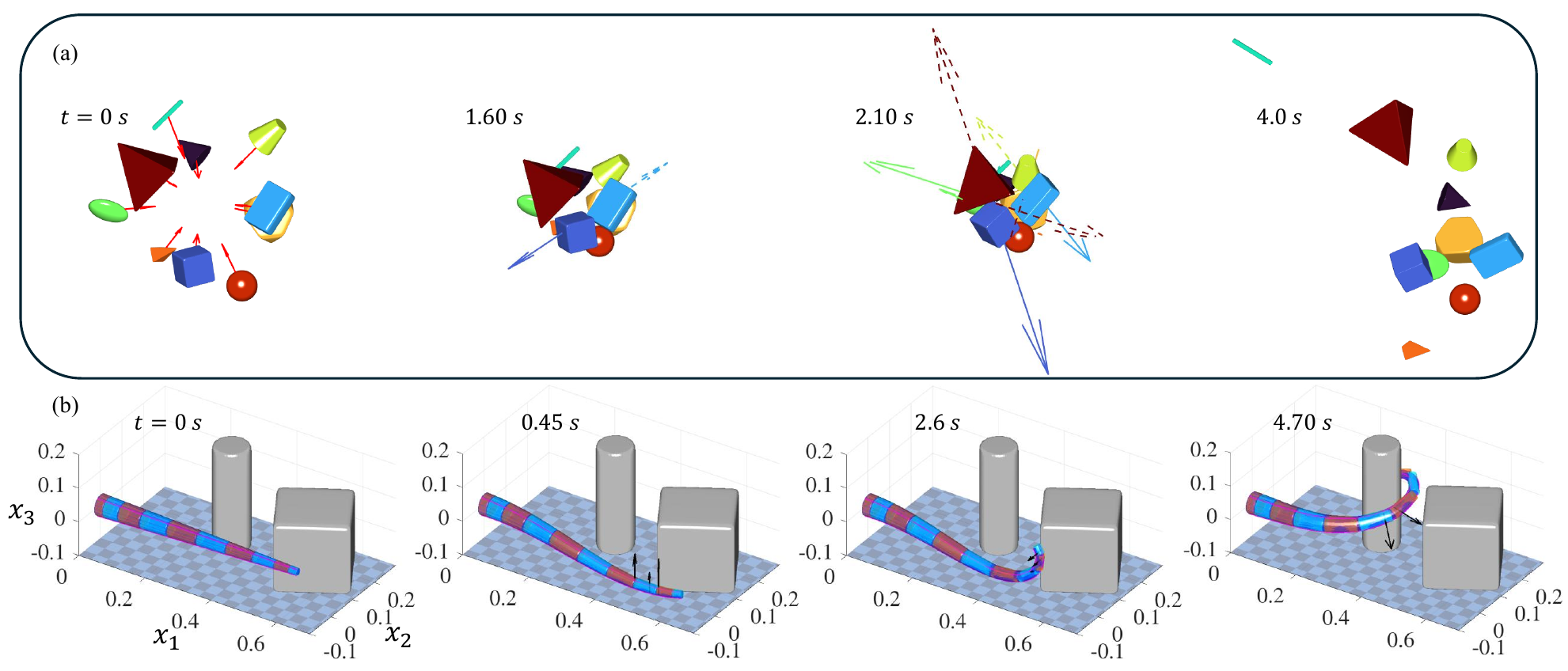}
    \caption{Contact physics examples: (a) Multibody rigid collisions among convex bodies: \emph{Cone, Cube, Cuboid, Cylinder, Ellipsoid, Frustum, Polytope, Pyramid, Sphere,} and \emph{Tetrahedron}. At $t=0$, bodies are thrown toward the origin with constant speed (red arrows). In subsequent snapshots, arrows and dashed arrows depict the action--reaction contact forces for each contact pair. (b) Interaction of a slender soft manipulator with rigid obstacles and the ground.}
    \label{fig:contact_physics}
\end{figure*}

We adopt a simple, frictionless,
penalty-based contact model (Hertz law) \cite{Flores2022}, in which the normal contact force is
\begin{equation}
\label{eq:contactmodel}
    f_n = k\,\delta^{p},
\end{equation}
where $\delta=-d_D$ denotes the penetration measure \eqref{eq:gap}, $k>0$ is the contact stiffness coefficient, and $p \ge 1$ controls the force growth rate.

The contact force is assumed to act at $\bm{x}^*$ in the direction opposite to $\hat{\bm{n}}(\bm{x}^*)$ \eqref{eq:unit_normal} in the local frame of body~1. The resulting contact wrench and its equal-and-opposite reaction applied to body~2 are given by
\begin{equation}
\label{eq:contact_wrench}
\bm{\mathcal{F}}_{c1} = -\begin{bmatrix}
\tilde{\bm{x}}^*\hat{\bm{n}}\\
\hat{\bm{n}}
\end{bmatrix} f_n,
\qquad
\bm{\mathcal{F}}_{c2} = -\mathrm{Ad}_{\bm{g}}^{-*}\bm{\mathcal{F}}_{c1}.
\end{equation}
where, $\mathrm{Ad}_{\bm{g}}^{*}$ denotes the co-Adjoint map in $SE(3)$ \cite{ModernRobotics}.

Jacobians of the contact wrench with respect to $\bm{q}$ (see Appendix) are incorporated into the equations of motion, yielding analytical derivatives of the dynamic model, consistent with prior formulations for rigid-body contact dynamics \cite{carpentier:hal-01790971,Singh2022} and hybrid soft–rigid robots \cite{mathew2026}. The approach is evaluated on rigid multibody and soft manipulator examples using the SoRoSim MATLAB toolbox \cite{Mathew2022}, which employs the geometric variable strain (GVS) formulation for hybrid soft–rigid robots, with soft links modeled as Cosserat rods \cite{Boyer_TRO2020}. Time integration is performed using MATLAB’s implicit \texttt{ode15s} solver with supplied analytical Jacobians. Contact resolution uses a broad-phase filter based on outer-sphere bounds and a warm-started solver for successive contact queries.

\subsubsection{Multibody Collision} 

Ten convex rigid bodies with heterogeneous geometries, constructed using the implicit families described in Sec.~\ref{sec:family}, are initialized with random orientations and distributed on a sphere of radius $R = 6$ units centered at the origin. Each body is assigned an initial velocity of constant magnitude directed toward the origin. Each rigid body has six DOF, parameterized by exponential coordinates, giving $n_{\text{dof}} = 60$. All pairwise interactions are considered, resulting in $45$ potential contact pairs. The $5$ s dynamic simulation is computed on average in approximately $1.94$ s, with narrow-phase collision detection accounting for $6.3\%$ of the total runtime. Representative snapshots of the dynamic simulation are shown in Fig.~\ref{fig:contact_physics}(a). Without analytical derivatives of the dynamic model, the same simulation requires approximately 6.5 times longer to compute on average and exhibits reduced numerical robustness.

\subsubsection{Soft Manipulator Interaction}
The second scenario considers a soft–rigid contact simulation involving a slender soft manipulator modeled using the GVS formulation. All angular Cosserat rod strains are enabled and parameterized using a fourth-order polynomial strain basis ($n_{\text{dof}} = 15$) \cite{mathew2026}. The manipulator is actuated via tendons with tensions linearly ramped from $0$ to $40 \; \text{N}$ over $5\;\text{s}$. The environment consists of a ground plane and two fixed rigid obstacles: a cube and a cylindrical block. For contact handling, the soft body is approximated by a set of smooth truncated cone primitives, each rigidly attached to a discrete cross section, as illustrated in Fig.~\ref{fig:contact_physics}(b).

The manipulator is initialized in a straight configuration with zero velocity. Under gravity, it first establishes contact with the ground, subsequently grazes the cube, and finally wraps around the cylindrical obstacle under sustained contact. Representative snapshots of the interaction are shown in Fig.~\ref{fig:contact_physics}(b). The differentiable simulation was completed in $5.64\;\text{s}$, with collision detection accounting for $6.4\%$ of the total runtime. Without analytical derivatives, \texttt{ode15s} stalled at intermediate simulation times, leading to runtimes that were up to two orders of magnitude longer.

\section{Discussions and Conclusion}
\label{sec:conclusions}

Contact kinematics exhibits a fundamental geometric limitation for differentiable collision models, arising from non-strictly convex geometry. By adopting strictly convex implicit shape representations, iDCOL mitigates these degeneracies by restoring local uniqueness and curvature of the contact solution. While sharper shapes (larger $\beta$ or $n$) recover the behavior of non-strictly convex geometry, moderate smoothing significantly improves the conditioning of the contact Jacobian at the cost of exact geometric fidelity.

Overall, the resulting advantages of iDCOL can be summarized as follows:

\begin{itemize}
  \item Employs tunable strictly convex implicit surface representations that regularize edges, corners, and flat regions, yielding locally unique and differentiable contact kinematics and a full-rank KKT Jacobian for moderate values of $\beta$ or $n$.

  \item The formulation yields a fixed-size KKT system with a deterministic $6\times6$ contact Jacobian $\bm{J}_c$ (or $\bm{T}_c$), in contrast to DCOL’s variable-structure conic programs, allowing analytical derivatives of contact kinematics to be computed efficiently.

  \item It enables the introduction of new convex primitives by specifying corresponding $\phi$, $\nabla\phi$, and $\nabla^2\phi$, while the remainder of the algorithm, including differentiation, is shape independent.
\end{itemize}

In summary, this work introduced iDCOL, a differentiable collision framework that addresses a fundamental tension between geometric exactness and analytical differentiability in contact-rich robotics. Building on the scaling-based perspective introduced in DCOL, iDCOL makes an explicit trade-off between exact geometric fidelity and strict convexity to mitigate geometric degeneracies that undermine differentiability. The resulting framework supports efficient microsecond-scale collision queries with analytical derivatives of contact kinematics, making it well-suited for gradient-based simulation and optimization. The iDCOL framework is publicly available \cite{iDCOLcode} and designed to be imported as a submodule within physics engines and trajectory optimization frameworks. We expect iDCOL to facilitate the development of scalable, contact-aware algorithms that more tightly couple geometry, dynamics, and optimization in robotics.

\section*{Acknowledgments}
This research was funded by the Center for Autonomous Robotic Systems, Khalifa University of Science and Technology 
(KU-CARS).

\appendices
\section{Derivatives of Implicit Shapes}
\label{app:shapes}

\setcounter{equation}{0}
\renewcommand{\theequation}{\Alph{section}\arabic{equation}}

\paragraph{Smooth Maximum}
Given scalar functions $\{\phi_i(\bm{y})\}_{i=1}^M$ and a smoothing parameter
$\beta>0$, define
\begin{equation}
m = \max_i \phi_i \,, \qquad
w_i = \frac{\exp\!\big(\beta(\phi_i-m)\big)}{\sum_j \exp\!\big(\beta(\phi_j-m)\big)} \, .
\end{equation}
The smooth maximum is
\begin{equation}
\phi(\bm{y}) = m + \frac{1}{\beta}\log\!\Big(\sum_{i=1}^M
\exp\!\big(\beta(\phi_i-m)\big)\Big) \, .
\end{equation}

We have,
\begin{align}
\nabla \phi &= \sum_{i=1}^M w_i\,\nabla \phi_i \, , \label{eq:smoothmax_grad} \\
\nabla^2 \phi &=
\sum_{i=1}^M w_i\,\nabla^2 \phi_i
+\beta\!\left(
\sum_{i=1}^M w_i\,\nabla \phi_i\nabla \phi_i^T
-(\nabla\phi)(\nabla\phi)^T
\right).
\label{eq:smoothmax_hessian}
\end{align}

\paragraph{Smooth Polytope}
Let $\bm{a}_i^T$ be the rows of $A\in\mathbb{R}^{m\times 3}$, $b_i\in\mathbb{R}$, and $L>0$.
\begin{equation}
\phi_i(\bm{y}) = \frac{\bm{a}_i^T\bm{y}-b_i}{L}, \qquad
\phi(\bm{y}) = \operatorname{smax}_\beta\{\phi_i(\bm{y})\}_{i=1}^m .
\end{equation}

Since each $\phi_i$ is affine, we have
\begin{equation}
\nabla \phi_i =  \bm{a}_i/L,
\qquad
\nabla^2 \phi_i = \bm{0}.
\end{equation}

Substituting $\nabla \phi_i$ and $\nabla^2 \phi_i$ into the \eqref{eq:smoothmax_grad} and \eqref{eq:smoothmax_hessian} yields
$\nabla \phi$ and $\nabla^2 \phi$.

\paragraph{Smooth Truncated Cone}
Let $(R_b,R_t,a,b)>0$ define the bottom radius, top radius, and half-heights,
with total height $h=a+b$. The radius varies linearly along the axial direction:
\begin{equation}
R(y_1) = R_b + (R_t - R_b)\frac{y_1 + a}{h}.
\end{equation}
Let $r^2 = y_2^2 + y_3^2$. The lateral surface and planar caps are described by
\begin{align}
\phi_s(\bm{y}) &= \frac{r^2}{R^2(y_1)} - 1, \\
\phi_b(\bm{y}) &= -\frac{y_1}{a} - 1, \\
\phi_t(\bm{y}) &= \frac{y_1}{b} - 1.
\end{align}
The smooth truncated cone is defined via a smooth maximum
\begin{equation}
\phi(\bm{y}) = \operatorname{smax}_\beta\{\phi_s,\phi_b,\phi_t\},
\end{equation}
with smoothing parameter $\beta>0$.

The gradients of the cap terms are constant,
\begin{equation}
\nabla \phi_b = \begin{bmatrix} -1/a & 0 & 0 \end{bmatrix}^T, \quad
\nabla \phi_t = \begin{bmatrix}  1/b & 0 & 0 \end{bmatrix}^T,
\end{equation}
with zero Hessians. Let $\kappa = \frac{R_t - R_b}{h}$. The gradient of the lateral term is
\begin{equation}
\nabla \phi_s =
\begin{bmatrix}
-2\,\kappa\,\dfrac{r^2}{R^3(y_1)} \\
\phantom{-}2\,\dfrac{y_2}{R^2(y_1)} \\
\phantom{-}2\,\dfrac{y_3}{R^2(y_1)}
\end{bmatrix},
\end{equation}
and its Hessian has the nonzero entries
\begin{align}
\frac{\partial^2 \phi_s}{\partial y_1^2} &=
6\,\kappa^2\,\frac{r^2}{R^4(y_1)}, \\
\frac{\partial^2 \phi_s}{\partial y_2^2} &=
\frac{2}{R^2(y_1)}, \qquad
\frac{\partial^2 \phi_s}{\partial y_3^2} =
\frac{2}{R^2(y_1)}, \\
\frac{\partial^2 \phi_s}{\partial y_1 \partial y_2} &=
-4\,\kappa\,\frac{y_2}{R^3(y_1)}, \quad
\frac{\partial^2 \phi_s}{\partial y_1 \partial y_3} =
-4\,\kappa\,\frac{y_3}{R^3(y_1)}.
\end{align}

Finally, the gradient and Hessian of $\phi$ are obtained by applying \eqref{eq:smoothmax_grad} and \eqref{eq:smoothmax_hessian}.

\paragraph{Superellipsoid}
Let $n\in\mathbb{N}$ and $(a,b,c)>0$. Define the normalized coordinates
\begin{equation}
\overline{y}_1=\frac{y_1}{a}, \qquad
\overline{y}_2=\frac{y_2}{b}, \qquad
\overline{y}_3=\frac{y_3}{c}.
\end{equation}
The implicit function is
\begin{equation}
S(\bm{y})=
\overline{y}_1^{\,2n}+
\overline{y}_2^{\,2n}+
\overline{y}_3^{\,2n},
\qquad
\phi(\bm{y})=\psi(S(\bm{y}))-1 ,
\end{equation}
where $\psi(S)=S^{1/(2n)}$. Its derivatives are
\begin{equation}
\psi'(S)=\frac{1}{2n}\,S^{\frac{1}{2n}-1},
\quad
\psi''(S)=\frac{1}{2n}\Big(\frac{1}{2n}-1\Big)S^{\frac{1}{2n}-2}.
\end{equation}

The gradient of $S$ is
\begin{equation}
\nabla S=
\begin{bmatrix}
\frac{2n}{a}\,\overline{y}_1^{\,2n-1} \\
\frac{2n}{b}\,\overline{y}_2^{\,2n-1} \\
\frac{2n}{c}\,\overline{y}_3^{\,2n-1}
\end{bmatrix},
\end{equation}
and the Hessian is diagonal,
\begin{equation}
\nabla^2 S=\operatorname{diag}(h_1,h_2,h_3),
\end{equation}
with
\begin{align}
h_1 &= \frac{2n(2n-1)}{a^2}\,\overline{y}_1^{\,2n-2},\\
h_2 &= \frac{2n(2n-1)}{b^2}\,\overline{y}_2^{\,2n-2},\\
h_3 &= \frac{2n(2n-1)}{c^2}\,\overline{y}_3^{\,2n-2}.
\end{align}

The gradient and Hessian of $\phi$ follow from the chain rule,
\begin{align}
\nabla \phi &= \psi'(S)\,\nabla S, \label{eq:phi_grad} \\
\nabla^2 \phi &= \psi'(S)\,\nabla^2 S
+ \psi''(S)\,(\nabla S)(\nabla S)^T . \label{eq:phi_hess}
\end{align}

For $n>1$, strict convexity is ensured in practice by a small regularization
\begin{equation}
\overline{y}_i^{\,2} \;\mapsto\; \overline{y}_i^{\,2}+\varepsilon, \qquad \varepsilon>0,
\end{equation}
which prevents vanishing curvature along coordinate planes.

\paragraph{Superelliptic Cylinder}
Let $n\in\mathbb{N}$ and $(R,h)>0$. Define
\begin{equation}
\overline{y}_1=\frac{y_1}{h}, \qquad
\overline{r}^2=\frac{y_2^2+y_3^2}{R^2}.
\end{equation}
The implicit function is
\begin{equation}
S(\bm{y})=\overline{y}_1^{\,2n}+\overline{r}^{\,2n},
\qquad
\phi(\bm{y})=\psi(S(\bm{y}))-1 ,
\end{equation}
using the same scalar map $\psi$ as defined above.

The gradient of $S$ is
\begin{align}
\frac{\partial S}{\partial y_1}
&= \frac{2n}{h}\,\overline{y}_1^{\,2n-1},\\
\frac{\partial S}{\partial y_2}
&= 2n\,\overline{r}^{\,2n-2}\,\frac{y_2}{R^2},\\
\frac{\partial S}{\partial y_3}
&= 2n\,\overline{r}^{\,2n-2}\,\frac{y_3}{R^2},
\end{align}
for $\overline{r}^2>0$.

The nonzero entries of the Hessian of $S$ are
\begin{align}
\frac{\partial^2 S}{\partial y_1^2}
&= \frac{2n(2n-1)}{h^2}\,\overline{y}_1^{\,2n-2},\\
\frac{\partial^2 S}{\partial y_2^2}
&= \frac{2n}{R^2}\,\overline{r}^{\,2n-4}
\Big((2n-1)y_2^2+y_3^2\Big),\\
\frac{\partial^2 S}{\partial y_3^2}
&= \frac{2n}{R^2}\,\overline{r}^{\,2n-4}
\Big((2n-1)y_3^2+y_2^2\Big),\\
\frac{\partial^2 S}{\partial y_2\partial y_3}
&= \frac{2n(2n-2)}{R^2}\,
\overline{r}^{\,2n-4}\,y_2 y_3 .
\end{align}

The gradient and Hessian of $\phi$ again follow from
\eqref{eq:phi_grad}–\eqref{eq:phi_hess}.

For $n>1$, we employ a small regularization
\begin{equation}
\overline{y}_1^{\,2} \;\mapsto\; \overline{y}_1^{\,2}+\varepsilon,
\qquad
\overline{r}^{\,2} \;\mapsto\; \overline{r}^{\,2}+\varepsilon,
\qquad \varepsilon>0,
\end{equation}
which restores strictly positive curvature in the radial directions and
improves numerical conditioning along the cylinder axis.

\section{Derivatives of Contact Kinematics and Wrenches}
\label{app:contact}

\paragraph{Derivatives of Contact Kinematics}

The iDCOL gap function $d_D$ depend on $\bm{q}$ through $\alpha^*$ and $\bm{r}$. Differentiating~\eqref{eq:gap} yields
\begin{equation}
\label{eq:ddD_dq}
\frac{\partial d_D}{\partial \bm{q}}
=
\frac{\|\bm{r}\|}{(\alpha^*)^{2}}\,
\frac{\partial \alpha^*}{\partial \bm{q}}
+
\left(1-\frac{1}{\alpha^*}\right)
\frac{\bm{r}^T}{\|\bm{r}\|}\,
\frac{\partial \bm{r}}{\partial \bm{q}} .
\end{equation}
where, $\frac{\partial \bm{r}}{\partial \bm{q}} = \bm{R}\bm{J}_v$, the translational part of the geometric Jacobian.

The derivatives of the witness points $\bm{p}_1$ and $\bm{p}_2$ defined in~\eqref{eq:witness} are obtained analogously by applying the product rule with respect to $\bm{x}^*$ and $\alpha^*$.

The unit normal $\hat{\bm{n}}(\bm{x}^)$ in the local frame of body~1, defined in~\eqref{eq:unit_normal}, depends on $\bm{q}$ only through $\bm{x}^*$. Its derivative is therefore
\begin{equation}
\label{eq:dnhat_dq}
\frac{\partial \hat{\bm{n}}}{\partial \bm{q}}
=
-\frac{1}{\|\nabla \phi\|}
\,\left(\tilde{\hat{\bm{n}}}\right)^{\,2}\,
\nabla^2 \phi\,
\frac{\partial \bm{x}^*}{\partial \bm{q}},
\end{equation}
where $\nabla \phi$ and $\nabla^2 \phi$ are evaluated at $\bm{x}^*$.

\paragraph{Derivatives of Contact Wrenches}
We use the penalty-based, frictionless normal model \eqref{eq:contactmodel} with $\delta=-d_D$, where $d_D$ is the iDCOL gap \eqref{eq:gap}. Hence,
\begin{equation}
\frac{\partial f_n}{\partial \bm{q}}
=
k\,p\,\delta^{p-1}\frac{\partial \delta}{\partial \bm{q}}
=
-\,k\,p\,\delta^{p-1}\frac{\partial d_D}{\partial \bm{q}},
\end{equation}
with $\frac{\partial d_D}{\partial \bm{q}}$ given in \eqref{eq:ddD_dq}. The contact force on body~1 is
$\bm{f}_n=-f_n\hat{\bm{n}}$, where $\hat{\bm{n}}$ is defined in \eqref{eq:unit_normal} and $\frac{\partial \hat{\bm{n}}}{\partial \bm{q}}$ is given by \eqref{eq:dnhat_dq}. Therefore,
\begin{equation}
\label{eq:dfc_dq}
\frac{\partial \bm{f}_n}{\partial \bm{q}}
=
-\,\hat{\bm{n}}\frac{\partial f_n}{\partial \bm{q}}
-\,f_n\frac{\partial \hat{\bm{n}}}{\partial \bm{q}}.
\end{equation}

The contact wrenches acting on the contact pairs, expressed in their body frames, are given by \eqref{eq:contact_wrench}. Their derivatives follow from the product rule:
\begin{equation}
\label{eq:dFc1_dq}
\frac{\partial \bm{\mathcal{F}}_{c1}}{\partial \bm{q}}
=
\begin{bmatrix}
-\tilde{\bm{f}}_n\,\frac{\partial \bm{x}^*}{\partial \bm{q}}
+\tilde{\bm{x}}^*\,\frac{\partial \bm{f}_n}{\partial \bm{q}}
\\[2pt]
\frac{\partial \bm{f}_n}{\partial \bm{q}}
\end{bmatrix},
\end{equation}

\begin{equation}
\label{eq:dFc2_dq}
\frac{\partial \bm{\mathcal{F}}_{c2}}{\partial \bm{q}}
=
-\mathrm{Ad}_{\bm{g}}^{-*}\frac{\partial \bm{\mathcal{F}}_{c1}}{\partial \bm{q}}
-
\overline{\mathrm{ad}}^*_{\bm{\mathcal{F}}_{c2}}
\bm{J}.
\end{equation}
where, for $\bm{\mathcal{V}} = [\bm{\mathsf{w}}^T\; \bm{\mathsf{v}}^T]^T \in \mathbb{R}^6$, the coadjoint-bar operator on $\mathfrak{se}(3)$ is given by

\begin{equation}
\overline{\mathrm{ad}}_{\bm{\mathcal{V}}}^* = -\left( \begin{array}{cc} \widetilde{\bm{\mathsf{w}}}& \widetilde{\bm{\mathsf{v}}} \\ \widetilde{\bm{\mathsf{v}}}& \bm{0}_{3\times3}\end{array} \right) \in \mathbb{R}^{6\times6} 
\end{equation}

\bibliographystyle{IEEEtran}
\bibliography{references}

@ARTICLE{Boyer_TRO2020,
author={F. {Boyer} and V. {Lebastard} and F. {Candelier} and F. {Renda}},
journal={IEEE Transactions on Robotics}, 
title={Dynamics of Continuum and Soft Robots: A Strain Parameterization Based Approach}, 
year={2020},
volume={},
number={},
pages={1-17},
doi={10.1109/TRO.2020.3036618}
}

@article{Mathew2022,
archivePrefix = {arXiv},
arxivId = {2107.05494},
author = {Mathew, Anup Teejo and Hmida, Ikhlas Mohamed Ben and Armanini, Costanza and Boyer, Frederic and Renda, Federico},
doi = {10.1109/MRA.2022.3202488},
eprint = {2107.05494},
issn = {1558223X},
journal = {IEEE Robotics and Automation Magazine},
title = {{SoRoSim: A MATLAB Toolbox for Hybrid Rigid-Soft Robots Based on the Geometric Variable-Strain Approach}},
year = {2022a}
}

@book{ModernRobotics,
author = {Lynch, Kevin M. and Park, Frank C.},
title = {Modern Robotics: Mechanics, Planning, and Control},
year = {2017},
isbn = {1107156300},
publisher = {Cambridge University Press},
address = {USA},
edition = {1st}
}

@inproceedings{Todorov2012MuJoCo,
   author = {Emanuel Todorov and Tom Erez and Yuval Tassa},
   doi = {10.1109/IROS.2012.6386109},
   isbn = {9781467317375},
   issn = {21530858},
   booktitle = {IEEE International Conference on Intelligent Robots and Systems},
   pages = {5026-5033},
   title = {MuJoCo: A physics engine for model-based control},
   year = {2012},
}

@inproceedings{Carpentier2019,
   author = {Justin Carpentier and Guilhem Saurel and Gabriele Buondonno and Joseph Mirabel and Florent Lamiraux and Olivier Stasse and Nicolas Mansard},
   doi = {10.1109/SII.2019.8700380},
   booktitle = {2019 IEEE/SICE International Symposium on System Integration (SII)},
   pages = {614-619},
   title = {The Pinocchio C++ library : A fast and flexible implementation of rigid body dynamics algorithms and their analytical derivatives},
   year = {2019},
}

@inproceedings{carpentier:hal-01790971,
  TITLE = {{Analytical Derivatives of Rigid Body Dynamics Algorithms}},
  AUTHOR = {Carpentier, Justin and Mansard, Nicolas},
  URL = {https://laas.hal.science/hal-01790971},
  BOOKTITLE = {{Robotics: Science and Systems (RSS 2018)}},
  ADDRESS = {Pittsburgh, United States},
  HAL_LOCAL_REFERENCE = {Rapport LAAS n{\textdegree} 18124},
  YEAR = {2018},
  MONTH = Jun,
  HAL_ID = {hal-01790971},
  HAL_VERSION = {v2},
}

@misc{howell2023dojodifferentiablephysicsengine,
      title={Dojo: A Differentiable Physics Engine for Robotics}, 
      author={Taylor A. Howell and Simon Le Cleac'h and Jan Brüdigam and J. Zico Kolter and Mac Schwager and Zachary Manchester},
      year={2023},
      eprint={2203.00806},
      archivePrefix={arXiv},
      primaryClass={cs.RO},
}

@article{Giftthaler2017,
   author = {Markus Giftthaler and Michael Neunert and Markus Stäuble and Marco Frigerio and Claudio Semini and Jonas Buchli},
   doi = {10.1080/01691864.2017.1395361},
   issn = {15685535},
   issue = {22},
   journal = {Advanced Robotics},
   month = {11},
   pages = {1225-1237},
   publisher = {Robotics Society of Japan},
   title = {Automatic differentiation of rigid body dynamics for optimal control and estimation},
   volume = {31},
   year = {2017},
}

@article{Singh2022,
   author = {Shubham Singh and Ryan P. Russell and Patrick M. Wensing},
   doi = {10.1109/LRA.2022.3141194},
   issn = {23773766},
   issue = {2},
   journal = {IEEE Robotics and Automation Letters},
   month = {4},
   pages = {1776-1783},
   publisher = {Institute of Electrical and Electronics Engineers Inc.},
   title = {Efficient Analytical Derivatives of Rigid-Body Dynamics Using Spatial Vector Algebra},
   volume = {7},
   year = {2022},
}

@article{Flores2022,
   author = {Paulo Flores},
   doi = {10.1007/s11044-021-09803-y},
   issn = {1573272X},
   issue = {2},
   journal = {Multibody System Dynamics},
   pages = {127-177},
   publisher = {The Author(s), under exclusive licence to Springer Nature B.V.},
   title = {Contact mechanics for dynamical systems: a comprehensive review},
   volume = {54},
   url = {http://dx.doi.org/10.1007/s11044-021-09803-y},
   year = {2022},
}

@INPROCEEDINGS{Tassa2014DDP,
  author={Tassa, Yuval and Mansard, Nicolas and Todorov, Emo},
  booktitle={2014 IEEE International Conference on Robotics and Automation (ICRA)}, 
  title={Control-limited differential dynamic programming}, 
  year={2014},
  volume={},
  number={},
  pages={1168-1175},
  doi={10.1109/ICRA.2014.6907001}}

@article{Newbury2024,
   author = {Rhys Newbury and Jack Collins and Kerry He and Jiahe Pan and Ingmar Posner and David Howard and Akansel Cosgun},
   doi = {10.1109/ACCESS.2024.3425448},
   issn = {21693536},
   journal = {IEEE Access},
   pages = {97581-97604},
   publisher = {Institute of Electrical and Electronics Engineers Inc.},
   title = {A Review of Differentiable Simulators},
   volume = {12},
   year = {2024}
}

@ARTICLE{GJK,
  author={Gilbert, E.G. and Johnson, D.W. and Keerthi, S.S.},
  journal={IEEE Journal on Robotics and Automation}, 
  title={A fast procedure for computing the distance between complex objects in three-dimensional space}, 
  year={1988},
  volume={4},
  number={2},
  pages={193-203},
  doi={10.1109/56.2083}}

@book{vanDenBergen2004,
  author    = {Gino van den Bergen},
  title     = {Collision Detection in Interactive 3D Environments},
  publisher = {Morgan Kaufmann},
  year      = {2004}
}

@INPROCEEDINGS{FCL,
  author={Pan, Jia and Chitta, Sachin and Manocha, Dinesh},
  booktitle={2012 IEEE International Conference on Robotics and Automation}, 
  title={FCL: A general purpose library for collision and proximity queries}, 
  year={2012},
  volume={},
  number={},
  pages={3859-3866},
  doi={10.1109/ICRA.2012.6225337}}

@article{Montaut2024,
   author = {Louis Montaut and Quentin Le Lidec and Vladimir Petrik and Josef Sivic and Justin Carpentier},
   doi = {10.1109/TRO.2024.3386370},
   issn = {19410468},
   journal = {IEEE Transactions on Robotics},
   pages = {2564-2581},
   publisher = {Institute of Electrical and Electronics Engineers Inc.},
   title = {GJK++: Leveraging Acceleration Methods for Faster Collision Detection},
   volume = {40},
   year = {2024}
}

@inproceedings{Montaut2023,
   author = {Louis Montaut and Quentin Le Lidec and Antoine Bambade and Vladimir Petrik and Josef Sivic and Justin Carpentier},
   doi = {10.1109/ICRA48891.2023.10160251},
   isbn = {9798350323658},
   issn = {10504729},
   booktitle = {Proceedings - IEEE International Conference on Robotics and Automation},
   pages = {3240-3246},
   publisher = {Institute of Electrical and Electronics Engineers Inc.},
   title = {Differentiable Collision Detection: a Randomized Smoothing Approach},
   volume = {2023-May},
   year = {2023}
}

@inproceedings{Zimmermann2022,
   author = {Simon Zimmermann and Matthias Busenhart and Simon Huber and Roi Poranne and Stelian Coros},
   doi = {10.1109/IROS47612.2022.9981093},
   issn = {21530866},
   booktitle = {IEEE International Conference on Intelligent Robots and Systems},
   pages = {8086-8093},
   publisher = {Institute of Electrical and Electronics Engineers Inc.},
   title = {Differentiable Collision Avoidance Using Collision Primitives},
   year = {2022}
}

@incollection{SDF,
  author    = {Curless, Brian and Levoy, Marc},
  title     = {A Volumetric Method for Building Complex Models from Range Images},
  booktitle = {Seminal Graphics Papers: Pushing the Boundaries, Volume 2},
  publisher = {Association for Computing Machinery},
  year      = {2023}
}

@inproceedings{Li2024,
  author    = {Yiming Li and Xuemin Chi and Amirreza Razmjoo and Sylvain Calinon},
  title     = {Configuration Space Distance Fields for Manipulation Planning},
  booktitle = {Robotics: Science and Systems},
  year      = {2024}
}

@inproceedings{Tracy2023,
   author = {Kevin Tracy and Taylor A. Howell and Zachary Manchester},
   doi = {10.1109/ICRA48891.2023.10160716},
   isbn = {9798350323658},
   issn = {10504729},
   booktitle = {Proceedings - IEEE International Conference on Robotics and Automation},
   pages = {3663-3670},
   publisher = {Institute of Electrical and Electronics Engineers Inc.},
   title = {Differentiable Collision Detection for a Set of Convex Primitives},
   volume = {2023-May},
   year = {2023}
}

@article{LeCleach2023,
   author = {Simon Le Cleac'h and Mac Schwager and Zachary Manchester and Vikas Sindhwani and Pete Florence and Sumeet Singh},
   doi = {10.1109/LRA.2023.3268824},
   issn = {23773766},
   issue = {7},
   journal = {IEEE Robotics and Automation Letters},
   month = {7},
   pages = {4012-4019},
   publisher = {Institute of Electrical and Electronics Engineers Inc.},
   title = {Single-Level Differentiable Contact Simulation},
   volume = {8},
   year = {2023}
}

@article{Escande2014,
   author = {Adrien Escande and Sylvain Miossec and Mehdi Benallegue and Abderrahmane Kheddar},
   doi = {10.1109/TRO.2013.2296332},
   issn = {15523098},
   issue = {3},
   journal = {IEEE Transactions on Robotics},
   pages = {666-678},
   publisher = {Institute of Electrical and Electronics Engineers Inc.},
   title = {A strictly convex hull for computing proximity distances with continuous gradients},
   volume = {30},
   year = {2014}
}

@misc{Lidec2025,
  author = {Quentin Le Lidec and Louis Montaut and Yann de Mont-Marin and Fabian Schramm and Justin Carpentier},
  title  = {End-to-End and Highly-Efficient Differentiable Simulation for Robotics},
  year   = {2025},
  note   = {arXiv preprint},
  url    = {https://arxiv.org/abs/2409.07107}
}

@misc{iDCOLcode,
  author       = {Anup Teejo Mathew},
  title        = {{iDCOL}: Implicit Differentiable Collision Detection},
  howpublished = {\url{https://github.com/SoRoSim/iDCOL}},
  year         = {2026},
  note         = {Software repository}
}

@misc{Jaitly2025,
  author = {Akshay Jaitly and Devesh K. Jha and Kei Ota and Yuki Shirai},
  title  = {Analytic Conditions for Differentiable Collision Detection in Trajectory Optimization},
  year   = {2025},
  note   = {arXiv preprint},
  url    = {http://arxiv.org/abs/2509.26459}
}

@INPROCEEDINGS{Ratliff2009,
  author={Ratliff, Nathan and Zucker, Matt and Bagnell, J. Andrew and Srinivasa, Siddhartha},
  booktitle={2009 IEEE International Conference on Robotics and Automation}, 
  title={CHOMP: Gradient optimization techniques for efficient motion planning}, 
  year={2009},
  volume={},
  number={},
  pages={489-494},
  doi={10.1109/ROBOT.2009.5152817}}

@article{Macklin2020,
author = {Macklin, Miles and Erleben, Kenny and M\"{u}ller, Matthias and Chentanez, Nuttapong and Jeschke, Stefan and Corse, Zach},
title = {Local Optimization for Robust Signed Distance Field Collision},
year = {2020},
issue_date = {Apr 2020},
publisher = {Association for Computing Machinery},
address = {New York, NY, USA},
volume = {3},
number = {1},
url = {https://doi.org/10.1145/3384538},
doi = {10.1145/3384538},
journal = {Proc. ACM Comput. Graph. Interact. Tech.},
month = may,
articleno = {8},
numpages = {17}
}

@article{CHEN20061053,
title = {Computing minimum distance between two implicit algebraic surfaces},
journal = {Computer-Aided Design},
volume = {38},
number = {10},
pages = {1053-1061},
year = {2006},
issn = {0010-4485},
doi = {https://doi.org/10.1016/j.cad.2006.04.012},
url = {https://www.sciencedirect.com/science/article/pii/S0010448506000765},
author = {Xiao-Diao Chen and Jun-Hai Yong and Guo-Qin Zheng and Jean-Claude Paul and Jia-Guang Sun}
}

@article{mathew2026,
author = {Anup Teejo Mathew and Frederic Boyer and Vincent Lebastard and Federico Renda},
title ={Analytical derivatives of strain-based dynamic model for hybrid soft-rigid robots},
journal = {The International Journal of Robotics Research},
volume = {45},
number = {1},
pages = {128-158},
year = {2026},
doi = {10.1177/02783649251346209},
eprint = { 
        https://doi.org/10.1177/02783649251346209
}
}

@article{lavalle2001randomized,
  title={Randomized kinodynamic planning},
  author={LaValle, Steven M and Kuffner Jr, James J},
  journal={The international journal of robotics research},
  volume={20},
  number={5},
  pages={378--400},
  year={2001},
  publisher={SAGE Publications}
}

@article{Nesterov2005,
  author  = {Nesterov, Yurii},
  title   = {Smooth Minimization of Non-Smooth Functions},
  journal = {Mathematical Programming},
  volume  = {103},
  number  = {1},
  pages   = {127--152},
  year    = {2005},
  doi     = {10.1007/s10107-004-0552-5}
}

@misc{DCOLcode,
  author       = {Kevin Tracy},
  title        = {DifferentiableCollisions.jl},
  howpublished = {\url{https://github.com/kevin-tracy/DifferentiableCollisions.jl}},
  year         = {2023},
  note         = {Julia package accompanying DCOL}
}

@article{Gonalves2017,
   author = {Artur Alves Gonçalves and Alexandre Bernardino and Joaquim Jorge and Daniel Simões Lopes},
   doi = {10.1016/j.mechmachtheory.2017.04.008},
   issn = {0094114X},
   journal = {Mechanism and Machine Theory},
   month = {9},
   pages = {77-96},
   publisher = {Elsevier Ltd},
   title = {A benchmark study on accuracy-controlled distance calculation between superellipsoid and superovoid contact geometries},
   volume = {115},
   year = {2017}
}

@article{Lopes2010,
   author = {Daniel S. Lopes and Miguel T. Silva and Jorge A. Ambrósio and Paulo Flores},
   doi = {10.1007/s11044-010-9220-0},
   issn = {13845640},
   issue = {3},
   journal = {Multibody System Dynamics},
   month = {10},
   pages = {255-280},
   title = {A mathematical framework for rigid contact detection between quadric and superquadric surfaces},
   volume = {24},
   year = {2010}
}

@ARTICLE{Barr1981,
  author={Barr},
  journal={IEEE Computer Graphics and Applications}, 
  title={Superquadrics and Angle-Preserving Transformations}, 
  year={1981},
  volume={1},
  number={1},
  pages={11-23},
  doi={10.1109/MCG.1981.1673799}}

@inproceedings{toussaint2018differentiable,
  title     = {Differentiable Physics and Stable Modes for Tool-Use and Manipulation Planning},
  author    = {Toussaint, Marc A. and Allen, Kelsey Rebecca and Smith, Kevin A. and Tenenbaum, Joshua B.},
  booktitle = {Proceedings of Robotics: Science and Systems (RSS)},
  year      = {2018}
}

@article{heess2015learning,
  title={Learning continuous control policies by stochastic value gradients},
  author={Heess, Nicolas and Wayne, Gregory and Silver, David and Lillicrap, Timothy and Erez, Tom and Tassa, Yuval},
  journal={Advances in neural information processing systems},
  volume={28},
  year={2015}
}

@inproceedings{carpentier2019pinocchio,
  title={The Pinocchio C++ library: A fast and flexible implementation of rigid body dynamics algorithms and their analytical derivatives},
  author={Carpentier, Justin and Saurel, Guilhem and Buondonno, Gabriele and Mirabel, Joseph and Lamiraux, Florent and Stasse, Olivier and Mansard, Nicolas},
  booktitle={2019 IEEE/SICE International Symposium on System Integration (SII)},
  pages={614--619},
  year={2019},
  organization={IEEE}
}

@article{bacher2021design,
  title={Design and control of soft robots using differentiable simulation},
  author={B{\"a}cher, Moritz and Knoop, Espen and Schumacher, Christian},
  journal={Current Robotics Reports},
  volume={2},
  number={2},
  pages={211--221},
  year={2021},
  publisher={Springer}
}

@INPROCEEDINGS{Strecke2021,
  author={Strecke, Michael and Stueckler, Joerg},
  booktitle={2021 International Conference on 3D Vision (3DV)}, 
  title={DiffSDFSim: Differentiable Rigid-Body Dynamics With Implicit Shapes}, 
  year={2021},
  volume={},
  number={},
  pages={96-105},
  doi={10.1109/3DV53792.2021.00020}}

@ARTICLE{LidecReview2024,
  author={Le Lidec, Quentin and Jallet, Wilson and Montaut, Louis and Laptev, Ivan and Schmid, Cordelia and Carpentier, Justin},
  journal={IEEE Transactions on Robotics}, 
  title={Contact Models in Robotics: A Comparative Analysis}, 
  year={2024},
  volume={40},
  number={},
  pages={3716-3733},
  doi={10.1109/TRO.2024.3434208}}

\newpage






\vfill

\end{document}